\documentclass{article}
\PassOptionsToPackage{numbers, sort&compress, authoryear}{natbib}

 \usepackage[main, final]{neurips_2026}

\usepackage[utf8]{inputenc} %
\usepackage[T1]{fontenc}    %
\usepackage{url}            %
\usepackage{booktabs}       %
\usepackage{amsfonts}       %
\usepackage{nicefrac}       %
\usepackage{microtype}      %
\usepackage{xcolor}         %

\definecolor{cvprblue}{rgb}{0.21,0.49,0.74}
\usepackage[pagebackref,breaklinks,colorlinks,allcolors=cvprblue]{hyperref}
\usepackage{cite}
\let\cite\citep
\usepackage{bbm}
\usepackage{graphicx}
\usepackage{amsmath}
\usepackage{amssymb}
\usepackage{amsmath}
\usepackage{duckuments}
\usepackage{algorithm}
\usepackage{algcompatible}
\usepackage{amsmath}
\usepackage{amsfonts}
\usepackage{amssymb}
\usepackage{mathtools}
\usepackage{float}
\usepackage[10pt]{moresize}
\usepackage{subcaption}
\usepackage{booktabs}
\usepackage{diagbox}

\usepackage{color}
\usepackage{epsfig}
\usepackage{graphicx}
\usepackage{times}

\usepackage{comment}
\usepackage{subcaption}
\usepackage{xcolor}
\usepackage{soul}
\usepackage{tabularx}
\newcolumntype{Y}{>{\centering\arraybackslash}X}

\usepackage{pifont}

\usepackage{microtype}
\usepackage[font=smaller]{caption}
\usepackage{tabularx}
\usepackage{adjustbox}
\usepackage{array}
\usepackage{booktabs}
\usepackage{colortbl}
\usepackage{float,wrapfig}
\usepackage{makecell}
\usepackage{hhline}
\usepackage{multirow}
\usepackage{subcaption} %
\usepackage[percent]{overpic}
\usepackage{graphbox}
\usepackage{diagbox}
\usepackage{stmaryrd}

\usepackage{amsmath,amsfonts,amssymb}
\usepackage{bm}
\usepackage{nicefrac}
\usepackage{microtype}
\usepackage{dsfont}
\usepackage{changepage}
\usepackage{extramarks}
\usepackage{fancyhdr}
\usepackage{setspace}
\usepackage{soul}
\usepackage{xspace}
\usepackage{hhline}

\usepackage{enumitem}
\usepackage[title]{appendix}

\usepackage{arydshln}

\usepackage{array}

\newcommand{\bx}{\mathbf{x}}
\newcommand{\by}{\mathbf{y}}
\newcommand{\bv}{\mathbf{V}}
\newcommand{\cs}{\mathcal{S}}

\newcommand{\R}[1]{{%
    \textbf{%
        \ifstrequal{#1}{1}{\textcolor{red}{R#1}}{%
        \ifstrequal{#1}{2}{\textcolor{blue}{R#1}}{%
        \ifstrequal{#1}{3}{\textcolor{magenta}{R#1}}{%
        \ifstrequal{#1}{4}{\textcolor{teal}{R#1}}{%
                           \textcolor{cyan}{R#1}%
        }}}}%
    }%
}}

\usepackage{amsmath,amsfonts,bm,relsize,dsfont}
\usepackage[numbers,sort&compress]{natbib}
\usepackage{anyfontsize}

\definecolor{MyDarkBlue}{rgb}{0.02,0.02,0.8}
\definecolor{MyDarkGreen}{rgb}{0.02,0.6,0.02}
\definecolor{MyDarkRed}{rgb}{0.8,0.02,0.02}
\definecolor{MyDarkOrange}{rgb}{0.40,0.2,0.02}
\definecolor{MyPurple}{RGB}{111,0,255}
\definecolor{ForestGreen}{RGB}{0, 128, 0}
\definecolor{MyRed}{rgb}{1.0,0.0,0.0}
\definecolor{MyGold}{rgb}{0.75,0.6,0.12}
\definecolor{MyDarkgray}{rgb}{0.66, 0.66, 0.66}
\definecolor{MyDarkCyan}{rgb}{0.05, 0.55, 0.45}
\definecolor{MyBlack}{rgb}{0., 0., 0.}
\definecolor{MyMagenta}{rgb}{1., 0., 1.}
\definecolor{BerkeleyYellow}{RGB}{255,204,41}
\definecolor{BerkeleyLightBlue}{RGB}{94,146,221}
\definecolor{BkDarkBlue}{rgb}{.05,.07,.353}

\definecolor{MyDarkGray2}{rgb}{0.6, 0.6, 0.6}

\definecolor{ClaudeOrange}{RGB}{204,85,0}

\newcommand{\ignorethis}[1]{}

\def\1{\bm{1}}

\newcommand{\ignore}[1]{}

\makeatletter
\DeclareRobustCommand\onedot{\futurelet\@let@token\@onedot}
\def\@onedot{\ifx\@let@token.\else.\null\fi\xspace}

\makeatother

\title{Dataset Distillation via Drifting}

\author{%
  George Cazenavette\hspace{4em}Angelina Quan\hspace{4em}Giannis Daras \\[6pt]
  \bf Antonio Torralba\hspace{4em}Vincent Sitzmann \\[6pt]
  Massachusetts Institute of Technology \\[6pt]
\mbox{\href{https://drifting-distillation.github.io}{drifting-distillation.github.io}}\\[6pt]
}

\begin{document}

\maketitle

\begin{center}
    \centering
    \vspace{-3em}
    \includegraphics[width=\linewidth]{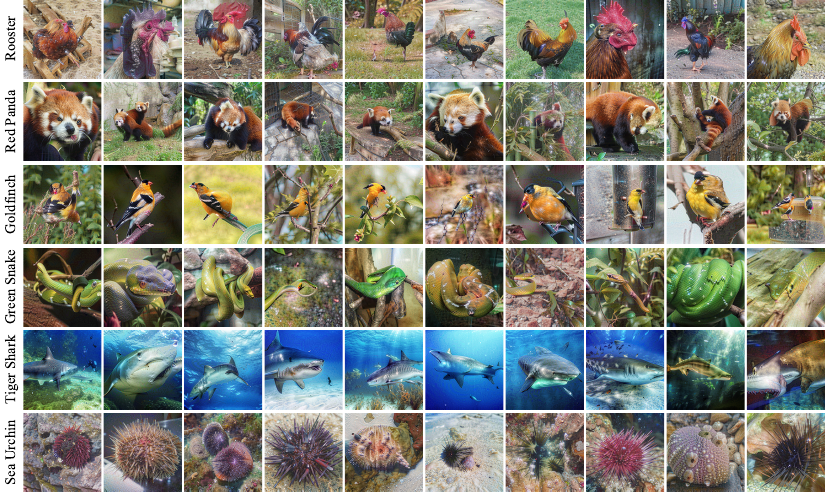}
    \captionof{figure}{\looseness=-1 
    A selection of 6 classes from ImageNet-1k~\cite{imagenet1k} distilled to 10 images per class via drifting with DINOv3~\cite{dinov3} features. Images seen here were initialized with \textit{random noise}. Our method yields synthetic data that are high in both quality and diversity by virtue of the attractive and repulsive forces of our drifting loss.
    }
    \label{fig:cover}
\end{center}

\begin{abstract}
\looseness=-1

The task of \textit{dataset distillation} concerns the learning of a small set of synthetic samples that can serve as an effective replacement for a large real dataset when used for downstream tasks. 
While many solutions have been proposed since the problem's inception, most of them primarily seek to optimize the distilled data with respect to a classifier. 
In this work, we propose a new distillation method that learns a set of samples such that they directly approximate the per-class data distribution. 
Specifically, we ``drift'' our distilled data in the latent space of a feature extractor such that they move both toward the real data and away from each other without using a classification objective. 
We find that our drifted data cover the real distribution significantly better than existing methods while still serving as powerful training samples for classification. 
Furthermore, we showcase our method's use as a visualization tool for better interpreting models and extend it beyond images to other modalities. 
Our work advances the frontier of dataset distillation, offering a novel synthesis method and proposing distribution-based metrics as a complementary evaluation. 
More broadly, our findings help address the growing need to understand both data and models by uncovering more about how they interact.
\end{abstract}

\section{Introduction}
\looseness=-1
Dataset distillation takes the classical problem of coreset selection~\cite{har2004coresets} (choosing a representative subset of a large dataset) and extends it by allowing for the \textit{learning} of a set of optimal samples rather than restricting to a selection of real data points. 
Since the task's proposal in the titular paper~\cite{dd}, dataset distillation solutions have focused on the optimization of synthetic samples for the purpose of training image classification models. 
Be it via gradient-~\cite{dc,dsa}, trajectory-~\cite{mtt,tesla,datm}, or feature-matching~\cite{dm,wang2022cafe,wmdd} or directly maximizing a classifier's confidence~\cite{sre2l,cda,gvbsm,scdd,cudd,fadrm,cvdd,rded}, distillation methods have traditionally indexed heavily toward the classification objective, often paying less attention to
other potentially desirable attributes of a tiny synthetic dataset. 
Data distilled with respect to a decision boundary inherit the classifier's biases, possibly limiting the data's usability for other tasks or generalization to other models.

\begin{figure}
    \centering
    \includegraphics[width=\linewidth]{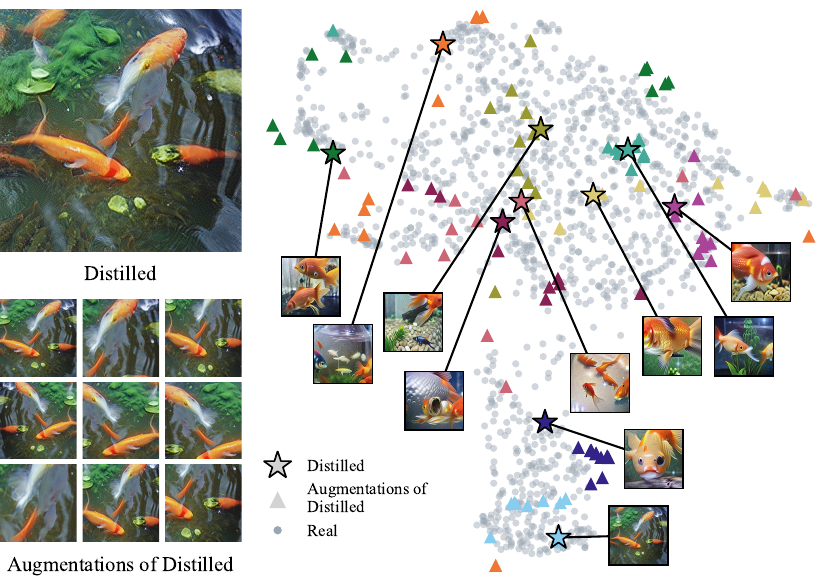}
    \caption{\looseness=-1\textbf{Augmentations of Distilled Images Cover the Real Distribution.} At each distillation step, we take multiple augmentations of each synthetic sample \textbf{(left)}. The synthetic data, together with their augmentations, approximately cover the distribution of real data \textbf{(right)}. In this plot, the stars represent the distilled images, the triangles 10 augmentations of each distilled image, and the circles real data. Drifting multiple augmentations of each synthetic sample at each step leads to high \textit{intra}-image diversity, allowing some distilled images (such as the \textcolor[HTML]{FF5733}{orange} star) to cover many parts of the real distribution at the same time.}
    \label{fig:scatter}
\end{figure}

\looseness=-1
In this work, we aim to distill a dataset with the explicit goal of \textbf{best approximating the real data distribution}. 
Inspired by recent work in generative modeling~\cite{drifting}, we learn our synthetic data via ``drifting'' such that each synthetic sample is attracted toward each real sample while simultaneously being repelled from every other synthetic sample.
The competing positive and negative forces encourage the distilled data to spread across the real distribution in a way that promotes both coverage and diversity. 
As a handful of images alone cannot adequately approximate the real distribution, we instead drift many \textbf{augmentations} of each synthetic image at a time such that the distilled images, together with their augmentations, best cover the target distribution (Figure~\ref{fig:scatter}). 

\looseness=-1
We find that this drifting method yields qualitatively more diverse and higher-quality images than existing methods (Figures~\ref{fig:cover} and~\ref{fig:noise_methods}). 
Despite being distilled without a classification objective, our images still serve as superior data for training such classifiers from scratch: the traditional dataset distillation benchmark (Sec.~\ref{sec:training}). 
Using MoCo-v3~\cite{mocov3} for the drifting feature space, our method reaches over 40\% and 52\% test accuracy when training a ResNet-18~\cite{resnet} from scratch with fixed soft labels after distilling ImageNet-1k~\cite{imagenet1k} down to just 10 and 50 images per class, respectively. 

\looseness=-1
We further evaluate the distilled data on a set of feature-space metrics (coverage~\cite{naeem2020reliable} along with nearest-neighbor and linear probe classification) where our method continues to match or outperform existing methods (Sec.~\ref{sec:distribution_metrics}).  
Our method also offers some model interpretability insights (Sec.~\ref{sec:interp}), generalizes to other modalities (Sec.~\ref{sec:audiovideo}), and even serves as an avenue for art (Sec.~\ref{sec:art}).

Code and distilled datasets are available on our project page: \mbox{\href{https://drifting-distillation.github.io}{drifting-distillation.github.io}}.

\section{Related Work}

\paragraph{Dataset Distillation.}
\looseness=-1
The concept of \textbf{coreset selection}~\citep{har2004coresets}, or finding a representative subset of a large dataset, has been thoroughly researched over many decades~\citep{feldman2011unified,welling2009herding,rebuffi2017icarl,toneva2018empirical,borsos2020coresets,mirzasoleiman2020coresets,sener2017active,paul2021deep,yang2024mind}. 
The idea of training a model to competitive performance using only a fraction of the data is attractive for many reasons, the most promising being the potential efficiency gains. 
More recently, \textbf{dataset distillation}~\citep{dd}, the \textit{learning} of a small, representative dataset, has gained favor, as the direct optimization of a set of data points allows for more expressive and content-rich examples than being constrained to a selection of real samples. 

\looseness=-1
For several years, \textit{gradient}- and \textit{trajectory}-matching methods~\citep{dd,dc,dsa,mtt,tesla,datm,cazenavette2025dataset} yielded the best performance on classification-based dataset distillation tasks. 
However, these expensive bi-level optimization methods fail to scale to realistic models (ResNet~\citep{resnet}, ViT~\citep{vit}, etc.) and datasets (e.g., ImageNet\nobreakdash-1k~\citep{imagenet1k}). 
In contrast, the relatively cheap, often task-agnostic, \textit{distribution}-matching methods~\citep{dm,wang2022cafe} scale computationally to large models and data but perform poorly at that scale.

Large-scale distillation began with the \textit{Squeeze, Recover, and Relabel}~\citep{sre2l} paradigm, which adopts the DeepInversion~\citep{deepinversion} objective: a DeepDream-style~\citep{deepdream} model inversion that adds BatchNorm-statistic matching to the cross-entropy minimization with respect to a pre-trained classifier. Many follow-up works have since refined this method~\citep{cda,gvbsm,edc,ufc,cudd,cvdd}. 
These methods were the first to perform competitively on realistic models and datasets, but the distilled data are effective mainly for \textit{knowledge distillation}~\citep{hinton2015distilling} rather than from-scratch training. 
Some methods address the homogeneity issues that come with the classification objective via diversity-promoting regularizers~\citep{dwa,delt,prism,dire}. 

A related line of work instead selects (and then optionally further optimizes) real image patches by scoring them with a pre-trained classifier~\citep{rded,nrr,infoutil,fadrm,reta}. 
These methods are either restricted to, or heavily biased toward, the empirical data distribution, severely limiting their application to other areas, such as privacy~\citep{dong2022privacy} or interpretability~\citep{cazenavette2025dataset}.

Another emerging area of dataset distillation focuses on leveraging pre-trained (most often diffusion-based) generative models to construct the synthetic data~\citep{minimax,d4m,manifoldgd,cao2,d3hr}. 
While previous works made use of generative priors~\citep{glad} as a form of regularization, these methods leverage diffusion models to effectively draw information-dense samples from the real data manifold.

This work leverages newly proposed ideas from generative modeling to more effectively cover the data distribution. 
Notably, our method strictly focuses on approximating the real distribution and thus does not require a classification objective. Rather than requiring an extra loss term, diversity in our method arises naturally from the repulsive term of the drifting objective. 
Our method can also be initialized with random noise and does not make use of a pre-trained generative model.

\paragraph{Drifting.}
\looseness=-1
As a recently proposed approach to generative modeling~\citep{drifting}, \textbf{drifting} upends the typical paradigm of diffusion~\citep{ho2020denoising,hoogeboom2023simple,sohl2015deep} and flow-matching~\citep{lipman2022flow} models by moving the generated samples toward the target distribution at \textit{training} time, removing the need for a costly denoising process at inference by natively enabling sampling in a single forward pass.
In the simplest terms, each generated sample is simultaneously \textit{pulled} toward every real sample and \textit{pushed} away from every other generated sample (weighted by a softmax kernel) such that the distribution of generated samples eventually matches that of the real samples.

We apply drifting to \textbf{dataset distillation}. 
Rather than back-propagating the gradients of the drifting loss into the weights of a generative model, we \textit{directly optimize} a discrete set of images such that these images, together with their augmentations, cover the target distribution. 
Aside from this change in parameterization, our only modification is to jointly drift many augmentations of each image at once, treating each augmentation as a unique image when computing negative distances. 
The rest of the drifting paradigm remains unchanged.

\section{Method}
In this section, we first formally motivate our distribution matching objective (Sec.~\ref{sec:distribution_matching}) before defining our drifting loss (Sec.~\ref{sec:drifting}) and pyramid parameterization (Sec.~\ref{sec:pyramid}).
\subsection{Distribution Matching}\label{sec:distribution_matching}
\looseness=-1
The goal of the original drifting formulation~\cite{drifting} (or most generative modeling frameworks) is to learn a process that maps from one distribution ($p_{\epsilon}$) to another ($p_{\text{data}}$). 
Typically, $p_\epsilon$ is an infinitely sampleable, continuous noise distribution, and either a single- (drifting~\cite{drifting}, GAN~\cite{goodfellow2020generative}) or multi-step (diffusion~\cite{ho2020denoising,sohl2015deep}, flow matching~\cite{lipman2022flow}) model $f$ is learned to map from $p_\epsilon$ to $p_\text{data}$ such that $p_\text{data}\approx f_\#p_\epsilon$, where ``$f_\#$'' denotes the pushforward induced by the model. 

Rather than a generative model $f$, our goal is to \textit{directly learn} a small set of synthetic samples $\bx\in\mathcal{S}$ such that the distribution of these samples ($p_\text{syn}$) matches that of the real data: 
\begin{equation}
    p_\text{syn}\triangleq\frac{1}{|\mathcal{S}|}\sum_{\bx\in\mathcal{S}}\delta_\bx\approx p_\text{data}
\end{equation}
where $\delta$ is the Dirac delta distribution. 
As described, this is not feasible: a handful of synthetic samples alone cannot adequately approximate an entire distribution of real data. 
Rather than attempting to optimize our synthetic data such that they alone cover the real distribution, we sample from a distribution of \textit{augmentations} $A\sim P_A$ and optimize such that our samples, together with their augmentations, properly approximate the real distribution:
\begin{equation}
    \mathbb{E}_{A\sim P_A}[A_\#p_{\text{syn}}] \approx p_\text{data}
\end{equation}
In practice, $P_A$ consists of random resized crops, horizontal flips, and Gaussian noise. 
\subsection{Drifting for Dataset Distillation}\label{sec:drifting}
\looseness=-1
To match distributions, we adopt the \textbf{drifting field} of~\cite{drifting}. Rather than the ``evolution $\approx$ denoising'' analogy, we view it as a ``force'' field wherein the synthetic samples are pulled and pushed in various directions.
That is, for a given sample $\bx\in\mathcal{S}$ and drifting field $\mathbf{V}$, $\mathbf{V}(\bx)$ gives the total ``force'' acting on $\bx$. 
This leads to the following loss function where $\bx$ is ``pushed'' toward its drifted target $\bx+\mathbf{V}(\bx)$:
\begin{equation}\label{loss}
    \mathcal{L}
    = \sum_{\bx_i \in \mathcal{S}}
    \left\|
        \bx_i -
        \texttt{stopgrad}\left(\bx_i + \mathbf{V}(\bx_i)\right)
    \right\|_2^2 .
\end{equation}
The \texttt{stopgrad} operation provides a frozen target for the distilled samples. 
Rather than the weights of a generative model, this drifting loss is back-propagated \textbf{directly to our synthetic data}. 
Our drifting field $\bv$ consists of two components: positive (attractive) and negative (repulsive). 
For a given synthetic sample $\bx_i \in \cs$, we have:
\begin{equation}\label{eq:posneg}
\begin{aligned}
    \bv^+(\bx_i)&\coloneq \frac{1}{Z_\text{data}}\mathbb{E}_{\by\sim p_\text{data}}\left[\kappa(\bx_i, \by)(\by-\bx_i)\right]\\
        \bv^-(\bx_i)&\coloneq \frac{1}{Z_\text{syn}}\frac{1}{\left|\cs\right|}\sum_
        {j, j\not=i}
        \kappa(\bx_i,\bx_j)(\bx_j-\bx_i)
    \end{aligned}
\end{equation}
where $Z_\text{data}$ and $Z_\text{syn}$ are the normalization factors
\begin{equation}\label{eq:normalization}
    \begin{aligned}
    Z_\text{data} \coloneq \mathbb{E}_{y\sim p_\text{data}}\left[\kappa(\bx_i, \by)\right],\;\;\;\;\;\;\;\;
    Z_\text{syn} \coloneq \frac{1}{\left|\cs\right|}\sum_
        {j, j\not=i}
        \kappa(\bx_i,\bx_j)
    \end{aligned}
\end{equation}
and $\kappa$ is the kernel
\begin{equation}
    \kappa(\bx, \by) = \exp\left(-\frac{1}{\tau}\|\bx - \by\|_2\right)
\end{equation}
with temperature $\tau$. 
The combined drifting field can be written as 
\begin{equation}\label{eq:v}
    \bv(\bx)\coloneq \bv^+(\bx) - \bv^-(\bx)
\end{equation}
Intuitively, these positive and negative fields can be thought of as the attractive (or repulsive) force between a sample $\bx_i$ and all other real (or synthetic) samples weighted by the distances between them via a softmax operation. 
In short, each synthetic sample is optimized to both be close to real samples and far from other synthetic samples. 
A given synthetic sample $\bx_i$ will reach equilibrium when the total attractive and repulsive forces induced by the other real and synthetic samples cancel each other out, leading to $\bv(\bx_i)=\mathbf{0}$.

\looseness=-1
Rather than in pixel space, we compute the drifting field in the latent space of a pre-trained feature extractor, $\phi$. 
The choice of $\phi$ plays a crucial role, as it defines the space in which we want to match the real distribution; a stronger feature extractor may lead to better semantic alignment in image space. 
Combined with the augmentations $A\sim P_A$ described in Sec.~\ref{sec:distribution_matching}, we have our full loss function\footnote{Omitted here for readability: we additionally apply the same normalization terms as~\cite{drifting} (Appendix~\ref{sec:method_details}).}:
\begin{equation}\label{eq:loss}
    \mathcal{L} = \mathbb{E}_{A\sim P_A}\left[\sum_{\bx\in\mathcal{S}}\left\|\phi\left(A(\bx)\right) - \texttt{stopgrad}\left(\phi\left(A(\bx)\right) + \mathbf{V}(\phi\left(A(\bx)\right))\right)\right\|_2^2\right]
\end{equation}
In practice, multiple augmentations of the same synthetic sample are also considered \textit{negative} samples with respect to each other, leading to increased intra-\textit{sample} diversity as well as the inherent intra-class variety afforded by the drifting paradigm. 
We visualize our distilled distribution in Figure~\ref{fig:scatter} and include pseudocode of the full process in Algorithms~\ref{alg:1} and~\ref{alg:2}.

\subsection{Implicit Regularization via Laplacian Pyramids}\label{sec:pyramid}
\begin{figure}
    \centering
    \includegraphics[width=1.0\linewidth]{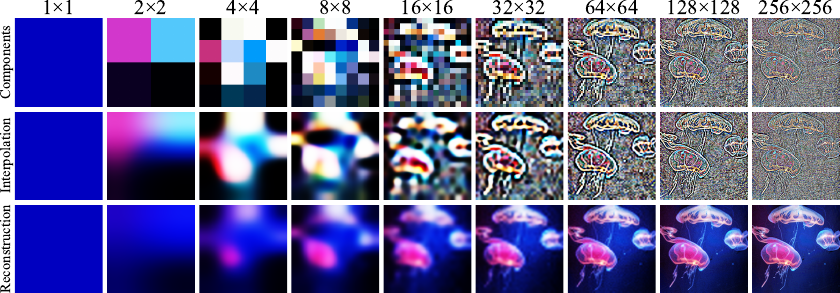}
    \caption{\looseness=-1 We represent distilled data as a Laplacian pyramid of cascading resolutions. To reconstruct a synthetic sample, each level (\textbf{top}) is bicubically interpolated to the full resolution (\textbf{middle}) before being added together \textbf{(bottom)}. This acts as a hyperparameter-free implicit regularizer on the distilled data, preventing overfitting to adversarial patterns often seen in model inversion tasks. 
    }
    \label{fig:pyramid}
\end{figure}

Inspired by recent work in dataset distillation~\cite{cazenavette2025dataset} and model inversion~\cite{fort2025direct}, we choose to parameterize our distilled data as a \textit{Laplacian pyramid} of different resolutions. 
Rather than naively optimizing the \textit{pixels} of an image, we instead represent each sample as a cascade of images with progressively lower resolutions: $\rho=\{1\times1, 2\times2,4\times4, \hdots, R\times R\}$ where $R\times R$ is the full resolution. 
Together, these layers form a Laplacian pyramid $P=\{\mathbf{P}_r|r\in\rho\}$.

At each optimization step, the synthetic sample $\bx$ is first built in the same manner an image is reconstructed from its Laplacian pyramid; each layer of the pyramid is bicubically upsampled to the canonical resolution before all layers are added together:
\begin{equation}\label{eq:pyramid}
    \bx = \operatorname{sigmoid}\left(\sum_{r\in\rho}\operatorname{resize}_{R}\left(\mathbf{P}_r\right)\right)
\end{equation}
\looseness=-1
The sigmoid operation acts as a soft, differentiable clamp to keep the range of the synthetic samples within that of the real data. 
As with prior work~\cite{cazenavette2025dataset}, we initialize our pyramid at low resolution and progressively add more layers throughout the distillation process. 
We also apply a fixed 3$\times$3 decorrelation transform to the color channels such that the distilled images are biased toward natural colors~\cite{olah2017feature}.

Notably, this pyramid technique generalizes effortlessly to both 1D (audio) and 3D (video) data, as seen in Section~\ref{sec:audiovideo}. 
Please see Figure~\ref{fig:pyramid} for a visualization of one of our distilled image pyramids.

\section{Experiments}
\begin{table}[t]
\centering
\small
\begin{tabular}{l cc cc cc}
\toprule
 & \multicolumn{2}{c}{Hard-Label} & \multicolumn{2}{c}{Soft-Label} & \multicolumn{2}{c}{Knowledge Distillation} \\
\cmidrule(lr){2-3}\cmidrule(lr){4-5}\cmidrule(lr){6-7}
Method & IPC 10 & IPC 50 & IPC 10 & IPC 50 & IPC 10 & IPC 50 \\
\midrule
\multicolumn{7}{l}{\emph{Noise init}} \\
SRe\textsuperscript{2}L~\cite{sre2l} & \phantom{0}8.1 \textcolor{gray}{$\pm$ 0.2} & 10.7 \textcolor{gray}{$\pm$ 0.3} & 17.5 \textcolor{gray}{$\pm$ 0.2} & 22.9 \textcolor{gray}{$\pm$ 0.3} & 40.6 \textcolor{gray}{$\pm$ 0.3} & 50.1 \textcolor{gray}{$\pm$ 0.1} \\
CDA~\cite{cda} & \phantom{0}9.5 \textcolor{gray}{$\pm$ 0.1} & 16.1 \textcolor{gray}{$\pm$ 0.3} & 19.2 \textcolor{gray}{$\pm$ 0.2} & 28.8 \textcolor{gray}{$\pm$ 0.1} & 41.1 \textcolor{gray}{$\pm$ 0.2} & 50.8 \textcolor{gray}{$\pm$ 0.1} \\
G-VBSM$^\dagger$~\cite{gvbsm} & 11.7 \textcolor{gray}{$\pm$ 0.3} & 24.9 \textcolor{gray}{$\pm$ 0.9} & 29.8 \textcolor{gray}{$\pm$ 0.2} & 44.9 \textcolor{gray}{$\pm$ 0.3} & 43.8 \textcolor{gray}{$\pm$ 0.3} & 53.0 \textcolor{gray}{$\pm$ 0.3} \\
WMDD~\cite{wmdd} & 10.2 \textcolor{gray}{$\pm$ 0.3} & 11.6 \textcolor{gray}{$\pm$ 0.6} & 29.4 \textcolor{gray}{$\pm$ 0.5} & 40.6 \textcolor{gray}{$\pm$ 0.1} & 43.7 \textcolor{gray}{$\pm$ 0.1} & 53.2 \textcolor{gray}{$\pm$ 0.2} \\
SC-DD~\cite{scdd} & \phantom{0}6.1 \textcolor{gray}{$\pm$ 0.2} & 10.2 \textcolor{gray}{$\pm$ 0.1} & 19.1 \textcolor{gray}{$\pm$ 0.2} & 31.0 \textcolor{gray}{$\pm$ 0.3} & 40.8 \textcolor{gray}{$\pm$ 0.2} & 51.2 \textcolor{gray}{$\pm$ 0.2} \\
Ours (ResNet-18) & 17.6 \textcolor{gray}{$\pm$ 0.2} & 18.4 \textcolor{gray}{$\pm$ 0.5} & 30.5 \textcolor{gray}{$\pm$ 0.8} & 43.1 \textcolor{gray}{$\pm$ 0.3} & \underline{44.8} \textcolor{gray}{$\pm$ 0.7} & 55.0 \textcolor{gray}{$\pm$ 0.3} \\
Ours (MoCo-v3) & \textbf{27.0} \textcolor{gray}{$\pm$ 0.2} & \textbf{33.5} \textcolor{gray}{$\pm$ 1.1} & \textbf{40.3} \textcolor{gray}{$\pm$ 0.1} & \textbf{52.8} \textcolor{gray}{$\pm$ 0.0} & \textbf{48.5} \textcolor{gray}{$\pm$ 0.1} & \textbf{57.2} \textcolor{gray}{$\pm$ 0.3} \\
Ours (DINOv3) & \underline{21.3} \textcolor{gray}{$\pm$ 0.5} & \underline{30.2} \textcolor{gray}{$\pm$ 0.2} & \underline{35.9} \textcolor{gray}{$\pm$ 0.1} & \underline{50.7} \textcolor{gray}{$\pm$ 0.1} & \underline{45.0} \textcolor{gray}{$\pm$ 0.2} & \underline{56.0} \textcolor{gray}{$\pm$ 0.2} \\
\midrule
\multicolumn{7}{l}{\emph{Real init}} \\
RDED~\cite{rded} & 22.1 \textcolor{gray}{$\pm$ 0.1} & 33.8 \textcolor{gray}{$\pm$ 1.0} & 33.1 \textcolor{gray}{$\pm$ 0.2} & 44.0 \textcolor{gray}{$\pm$ 0.3} & 46.7 \textcolor{gray}{$\pm$ 0.3} & 56.1 \textcolor{gray}{$\pm$ 0.1} \\
CUDD~\cite{cudd} & 15.2 \textcolor{gray}{$\pm$ 0.2} & 19.8 \textcolor{gray}{$\pm$ 0.4} & 28.3 \textcolor{gray}{$\pm$ 0.1} & 41.1 \textcolor{gray}{$\pm$ 0.3} & 46.6 \textcolor{gray}{$\pm$ 0.1} & 55.4 \textcolor{gray}{$\pm$ 0.1} \\
FADRM~\cite{fadrm} & 18.7 \textcolor{gray}{$\pm$ 0.1} & 25.7 \textcolor{gray}{$\pm$ 0.5} & 29.7 \textcolor{gray}{$\pm$ 0.1} & 40.0 \textcolor{gray}{$\pm$ 0.3} & 48.5 \textcolor{gray}{$\pm$ 0.3} & 56.4 \textcolor{gray}{$\pm$ 0.1} \\
FADRM+$^\dagger$~\cite{fadrm} & 20.5 \textcolor{gray}{$\pm$ 0.5} & 26.2 \textcolor{gray}{$\pm$ 0.6} & 32.5 \textcolor{gray}{$\pm$ 0.2} & 41.8 \textcolor{gray}{$\pm$ 0.3} & \textbf{49.2} \textcolor{gray}{$\pm$ 0.5} & 56.6 \textcolor{gray}{$\pm$ 0.1} \\
CV-DD$^\dagger$~\cite{cvdd} & 16.7 \textcolor{gray}{$\pm$ 0.0} & 21.9 \textcolor{gray}{$\pm$ 0.3} & 28.7 \textcolor{gray}{$\pm$ 0.3} & 38.3 \textcolor{gray}{$\pm$ 0.1} & 45.6 \textcolor{gray}{$\pm$ 0.1} & 54.2 \textcolor{gray}{$\pm$ 0.1} \\
Ours (ResNet-18) & 20.4 \textcolor{gray}{$\pm$ 0.4} & 22.0 \textcolor{gray}{$\pm$ 0.4} & 33.9 \textcolor{gray}{$\pm$ 0.2} & 45.4 \textcolor{gray}{$\pm$ 0.5} & 47.8 \textcolor{gray}{$\pm$ 0.2} & 56.4 \textcolor{gray}{$\pm$ 0.3} \\
Ours (MoCo-v3) & \textbf{28.9} \textcolor{gray}{$\pm$ 0.2} & \underline{36.0} \textcolor{gray}{$\pm$ 0.7} & \textbf{41.7} \textcolor{gray}{$\pm$ 0.0} & \underline{53.4} \textcolor{gray}{$\pm$ 0.1} & \textbf{49.6} \textcolor{gray}{$\pm$ 0.2} & \underline{57.7} \textcolor{gray}{$\pm$ 0.2} \\
Ours (DINOv3) & \underline{26.0} \textcolor{gray}{$\pm$ 0.2} & \textbf{41.1} \textcolor{gray}{$\pm$ 0.4} & \underline{40.5} \textcolor{gray}{$\pm$ 0.0} & \textbf{54.3} \textcolor{gray}{$\pm$ 0.4} & \textbf{49.5} \textcolor{gray}{$\pm$ 0.3} & \textbf{58.1} \textcolor{gray}{$\pm$ 0.1} \\
\bottomrule
\end{tabular}

\vspace{0.5em}
\caption{\looseness=-1 
\textbf{From-Scratch Training.} This table shows ImageNet-1k~\cite{imagenet1k} classification performance when training a ResNet-18~\cite{resnet} from scratch. We include results using one-hot hard labels (\textbf{left}), one-time soft labels (\textbf{middle}), and full knowledge distillation (\textbf{right}). Methods marked with a dagger$^\dagger$ distill using an ensemble of encoders. Across all three training paradigms, our distilled data lead to the best downstream performance. }
\label{tab:from_scratch}\vspace{-0.8em}\end{table}

As the focus of this work is \textit{large-scale} dataset distillation, our experimental results center on \mbox{ImageNet-1k}~\cite{imagenet1k}.
All distillation experiments are conducted one class at a time, allowing for easy parallelization across many GPUs. 
We primarily use three backbones for our image distillation experiments: ResNet-18~\cite{resnet}, DINOv3~\cite{dinov3}, and MoCo-v3~\cite{mocov3} (ViT-B for both transformers). The specific features to which we apply the drifting loss are detailed in Appendix~\ref{sec:method_details}.
\begin{figure}[!t]
    \centering
    \input{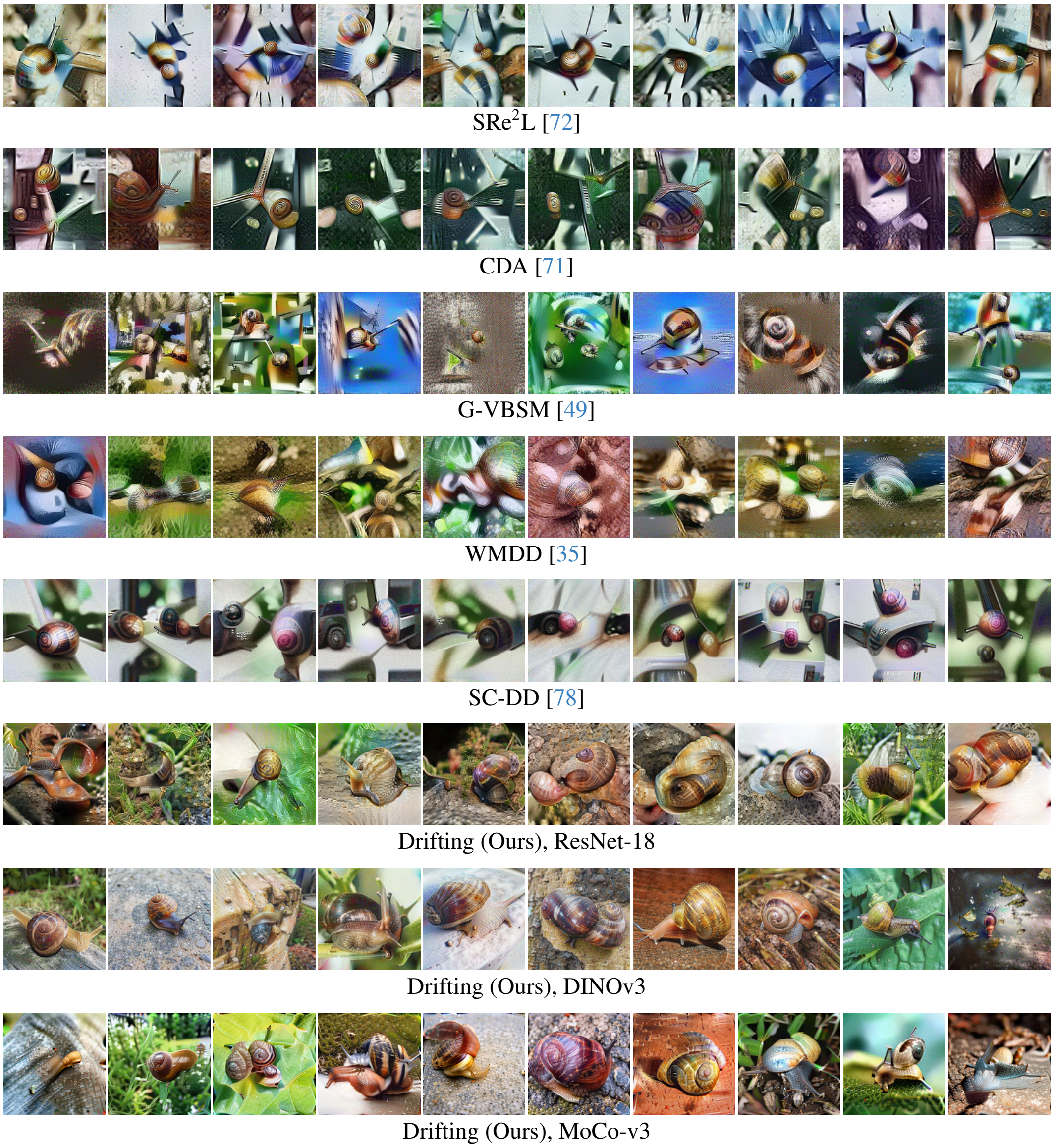}
    \vspace{-0.5em}
    \caption{\looseness=-1 \textbf{Comparing Noise-Initialized Distillation Methods.}  Even when using ResNet-18 as the distillation backbone, our method yields higher-quality and more diverse samples. When using a stronger backbone \mbox{(DINOv3~\cite{dinov3})}, the visual quality and diversity of our method increase further.
    }
    \label{fig:noise_methods}
    \vspace{-1em}
\end{figure}

\looseness=-1
We compare against both noise-initialized methods (SRe\textsuperscript{2}L~\cite{sre2l}, CDA~\cite{cda}, G-VBSM~\cite{gvbsm}, WMDD~\cite{wmdd}, and SC-DD~\cite{scdd}) and real-initialized methods (RDED~\cite{rded}, CUDD~\cite{cudd}, FADRM and FADRM+~\cite{fadrm}, and CV-DD~\cite{cvdd}). 
Since much of the improvement among real-initialized methods seems to come from how the initial images are chosen~\cite{rded,cudd,fadrm}, we report the two groups separately and run our own method under both initializations.
Noise initialization remains of particular interest, as initializing with real images limits usefulness for some downstream tasks, such as those reliant on privacy~\cite{dong2022privacy}. 
We compare against diffusion-based methods~\cite{minimax,d4m,d3hr,cao2,manifoldgd} in Appendix Table~\ref{tab:diffusion_compare}.
\subsection{Dataset Distillation via Drifting Yields Superior Training Samples.}\label{sec:training}
\looseness=-1
While many prior large-scale dataset distillation works bake the classification task into the data by directly minimizing a classifier's loss~\cite{sre2l,cda,gvbsm,scdd,cudd,fadrm,cvdd,rded}, our drifting method is \textit{solely} designed to match the per-class \textbf{distribution} of the real data. 
In fact, our method does not require a classification objective at all; drifting can occur in the space of any feature extractor, regardless of training objective.
Nevertheless, we still find that our synthetic images serve as excellent data for training classifiers.

In Table~\ref{tab:from_scratch}, we train ResNet-18~\cite{resnet} from scratch using the distilled data from each method in three settings: hard-label, soft-label, and knowledge distillation. 
Hard-label uses one-hot labels, while soft-label first computes a single fixed target vector for each training sample using a teacher model. 
Knowledge distillation (the evaluation method used in most prior large-scale distillation works) computes a new target vector at every step by passing the current augmented version of the training image through the teacher model; we include it for comparison with prior work, but note that it is effectively a teacher-student transfer task rather than from-scratch training (see Appendix~\ref{sec:method_details}). 
To avoid confounding evaluation factors, metrics for all methods were recomputed using the exact same training recipes (Appendix~\ref{sec:eval_protocols}). We report the mean and standard deviation over 3 fixed seeds.

For the hard- and soft-label settings, our method consistently outperforms both the noise- and real-initialized baselines. 
In the knowledge distillation setting, we still match or exceed the baselines despite the smaller gap (due to the task's tendency to saturate). 
The performance difference is most dramatic in the hard-label setting, especially with noise initialization. 
At just 10 images per class, our best noise-initialized variant (MoCo-v3~\cite{mocov3}) reaches 27.0\% test accuracy while the best noise-initialized baseline (G-VBSM~\cite{gvbsm}) achieves only 11.7\%. With real initialization, our MoCo-v3 data reach 28.9\% versus 22.1\% for the best baseline (RDED~\cite{rded}), and at 50 images per class the gaps remain large (33.5\% vs.\ 24.9\% with noise initialization and 41.1\% vs.\ 33.8\% with real initialization).

Notably, the best training data come from the self-supervised backbones (MoCo-v3 and DINOv3), which were never trained with labels, while the data distilled with ResNet-18 (a supervised classifier with the same architecture as the student) is the weakest of our three in every setting. Even so, our ResNet-18 data still outperforms the individual baselines in 49 of 60 pairwise comparisons (28 of 30 with noise initialization), and outperforms every noise-initialized baseline in four of the six settings. This suggests that the efficacy of our data stems from matching the real distribution in a rich feature space rather than from alignment with the student architecture or the classification task. We also note that G-VBSM, FADRM+, and CV-DD (marked $\dagger$ in Tables~\ref{tab:from_scratch} and~\ref{tab:feature_space}) distill against an ensemble of backbones, whereas each of our variants uses a single encoder.

\looseness=-1
We include results for additional distillation backbones in Appendix Table~\ref{tab:from_scratch_all} and report from-scratch performance on the ImageNette and ImageWoof~\cite{imagenette} datasets in Appendix Tables~\ref{tab:fromscratch_imagenette} and~\ref{tab:fromscratch_imagewoof}.

\subsection{Drifted Synthetic Datasets Cover the Real Distribution.}\label{sec:distribution_metrics}
\begin{table}[t]
\centering
\small
\setlength{\tabcolsep}{3pt}%
\resizebox{\linewidth}{!}{%
\begin{tabular}{l *{4}{c} *{4}{c} *{4}{c}}
\toprule
 & \multicolumn{4}{c}{Coverage} & \multicolumn{4}{c}{k-NN} & \multicolumn{4}{c}{Linear} \\
\cmidrule(lr){2-5}\cmidrule(lr){6-9}\cmidrule(lr){10-13}
 & \multicolumn{2}{c}{SigLIP2} & \multicolumn{2}{c}{MoCo-v3} & \multicolumn{2}{c}{SigLIP2} & \multicolumn{2}{c}{MoCo-v3} & \multicolumn{2}{c}{SigLIP2} & \multicolumn{2}{c}{MoCo-v3} \\
\cmidrule(lr){2-3}\cmidrule(lr){4-5}\cmidrule(lr){6-7}\cmidrule(lr){8-9}\cmidrule(lr){10-11}\cmidrule(lr){12-13}
Method & {\smaller IPC 10} & {\smaller IPC 50} & {\smaller IPC 10} & {\smaller IPC 50} & {\smaller IPC 10} & {\smaller IPC 50} & {\smaller IPC 10} & {\smaller IPC 50} & {\smaller IPC 10} & {\smaller IPC 50} & {\smaller IPC 10} & {\smaller IPC 50} \\
\midrule
\multicolumn{13}{l}{\emph{Noise init}} \\
SRe\textsuperscript{2}L~\cite{sre2l} & \phantom{0}7.2 & 11.6 & \phantom{0}4.9 & \phantom{0}8.5 & 50.1 & 54.3 & 46.0 & 49.1 & 51.2 & 55.6 & 41.1 & 45.4 \\
CDA~\cite{cda} & \phantom{0}6.5 & 11.1 & \phantom{0}4.6 & \phantom{0}8.7 & 44.5 & 51.7 & 44.8 & 49.5 & 51.0 & 57.7 & 43.2 & 50.9 \\
G-VBSM$^\dagger$~\cite{gvbsm} & \phantom{0}5.7 & \phantom{0}8.6 & \phantom{0}5.3 & \phantom{0}8.4 & 41.3 & 45.9 & 39.7 & 42.5 & 62.3 & 70.5 & 47.1 & 61.8 \\
WMDD~\cite{wmdd} & \phantom{0}9.5 & 17.5 & \phantom{0}9.4 & 19.0 & 49.6 & 56.0 & 45.8 & 50.7 & 65.2 & 66.4 & 45.7 & 58.7 \\
SC-DD~\cite{scdd} & \phantom{0}9.7 & 16.1 & \phantom{0}9.3 & 16.8 & 48.5 & 53.3 & 49.1 & 51.1 & 52.7 & 58.0 & 48.7 & 52.8 \\
Ours (ResNet-18) & 19.7 & 38.4 & \underline{41.3} & \underline{66.3} & 54.3 & 61.2 & \underline{55.3} & \underline{58.7} & 69.6 & 71.2 & \underline{59.8} & \underline{66.5} \\
Ours (MoCo-v3) & \underline{61.2} & \underline{86.9} & (91.9) & (97.9) & \underline{64.3} & \underline{68.4} & (60.2) & (62.5) & \underline{74.6} & \underline{74.6} & (66.6) & (69.9) \\
Ours (DINOv3) & \textbf{77.0} & \textbf{93.2} & \textbf{83.4} & \textbf{96.3} & \textbf{69.9} & \textbf{72.5} & \textbf{61.1} & \textbf{63.2} & \textbf{77.5} & \textbf{78.6} & \textbf{66.3} & \textbf{70.3} \\
\midrule
\multicolumn{13}{l}{\emph{Real init}} \\
RDED~\cite{rded} & 23.3 & 34.6 & 31.8 & 49.2 & \textbf{70.2} & \underline{71.6} & \textbf{61.0} & \underline{62.1} & 74.4 & \underline{77.2} & \underline{63.5} & 66.6 \\
CUDD~\cite{cudd} & 11.2 & 15.1 & 10.5 & 17.1 & 56.3 & 58.0 & 52.1 & 53.9 & 63.2 & 63.5 & 53.9 & 58.9 \\
FADRM~\cite{fadrm} & 28.5 & 43.0 & 30.3 & 52.3 & 62.8 & 66.9 & 57.8 & 61.3 & 67.6 & 71.5 & 61.1 & 64.9 \\
FADRM+$^\dagger$~\cite{fadrm} & 32.1 & 48.0 & 36.5 & 59.7 & 64.3 & 68.1 & 58.5 & 61.8 & 69.2 & 73.1 & 61.8 & 65.4 \\
CV-DD$^\dagger$~\cite{cvdd} & 10.0 & 16.6 & \phantom{0}8.4 & 15.8 & 55.7 & 61.1 & 51.8 & 54.9 & 62.7 & 66.6 & 52.7 & 57.4 \\
Ours (ResNet-18) & 30.0 & 57.5 & \underline{54.9} & \underline{82.0} & 62.1 & 68.3 & 58.0 & 61.3 & 72.8 & 74.5 & 63.1 & \underline{67.8} \\
Ours (MoCo-v3) & \underline{60.6} & \underline{88.5} & (91.3) & (98.0) & 65.0 & 69.3 & (60.2) & (62.2) & \underline{75.0} & 75.8 & (66.4) & (70.0) \\
Ours (DINOv3) & \textbf{79.9} & \textbf{95.5} & \textbf{85.1} & \textbf{97.5} & \underline{69.9} & \textbf{72.8} & \underline{60.3} & \textbf{63.0} & \textbf{77.0} & \textbf{78.5} & \textbf{65.7} & \textbf{70.2} \\
\bottomrule
\end{tabular}
}

\vspace{0.5em}
\caption{\looseness=-1 
\textbf{Feature-Space Metrics.} Coverage (\textbf{left}), $k$-NN accuracy (\textbf{middle}), and linear probe accuracy (\textbf{right}) of distilled ImageNet-1k data in the SigLIP2~\cite{siglip2} and MoCo-v3~\cite{mocov3} latent spaces (see Appendix~\ref{sec:eval_protocols}). Parenthesized entries are self-evaluations and are not ranked.
Methods marked with a dagger$^\dagger$ distill using an ensemble of encoders.
Our method outperforms all baselines in every feature-space setting except for RDED~\cite{rded} on IPC 10 $k$-NN in both latent spaces. The performance gap is especially large on the coverage metric.
}
\label{tab:feature_space}\vspace{-0.8em}\end{table}

\looseness=-1
While the traditional goal of dataset distillation methods has been the training of classifiers, our primary motivation is to learn a small set of synthetic images that best approximates the real distribution. 
To measure this directly, we evaluate the distilled data in the latent space of pre-trained encoders using three metrics: \textbf{coverage}~\cite{naeem2020reliable}, 1-nearest-neighbor ($k$-NN with $k=1$) classification, and linear probe accuracy (details in Appendix~\ref{sec:eval_protocols}).
Coverage measures the fraction of real validation points $\by_j\in\mathcal{T}_\text{val}$ whose neighborhoods, defined by their $k$-th nearest-neighbor distances $\operatorname{NND}_k(\by_j)$ with respect to other real points, contain at least one synthetic point:
\begin{equation}\label{eq:coverage}
    \mbox{Coverage} = \frac{1}{M}\sum_{j=1}^{M}\mathbbm{1}\left[\exists\, i\; \mbox{s.t.}\; d\big(\phi(\bx_i), \phi(\by_j)\big) \leq \operatorname{NND}_k(\by_j)\right]
\end{equation}
\looseness=-1
where $d$ is the cosine distance. We use $k=5$ and report the average coverage over all classes. We also add 10 random crops of each image to the pool to widen the synthetic distribution.

\looseness=-1
We evaluate in the latent spaces of SigLIP2~\cite{siglip2} (language-supervised) and MoCo-v3~\cite{mocov3} (self-supervised). 
Cells in which our MoCo-v3 variant is scored in its own distillation space are shown in parentheses and excluded from the rankings. Such self-evaluation seems to strongly inflate coverage (91.9\% versus 83.4\% for our DINOv3 variant at 10 images per class with noise initialization), but not $k$-NN or linear accuracy, where our MoCo-v3 variant is roughly on par with our DINOv3 variant.

\looseness=-1
In Table~\ref{tab:feature_space}, our DINOv3 variant achieves the highest coverage in every setting (excluding self-evaluation); at 10 images per class, it covers 77.0\% (SigLIP2) and 83.4\% (MoCo-v3) of the real distribution, compared to at most 9.7\% for any noise-initialized and 36.5\% for any real-initialized baseline. 
Our ResNet-18 variant covers more of the distribution than every noise-initialized baseline in all settings, and more than every real-initialized baseline in all but one (SigLIP2 at 10 images per class, where FADRM+ reaches 32.1\% versus our 30.0\%).
Our DINOv3 variant is also the best on $k$-NN and linear probe accuracy in every setting except two $k$-NN cells at 10 images per class, where RDED is ahead by less than one point.
Notably, the gaps in coverage are far larger than those in $k$-NN or linear accuracy: classification metrics only require the distilled data to separate classes, whereas coverage rewards spanning the full extent of each class distribution, which is precisely what drifting optimizes for.
Qualitatively, we see our distilled images are more diverse than those of the baseline methods, likely aiding in the distribution coverage (Figure~\ref{fig:noise_methods}).

Much of this distributional coverage comes from the distillation-time augmentations (Table~\ref{tab:abl_crops}), which allow our method to explicitly optimize for intra-image diversity.
We visualize this in Figure~\ref{fig:scatter}; while the distilled images themselves (stars) are relatively evenly spaced throughout the latent space, their augmentations (triangles) help cover the rest of the real distribution.

\subsection{Distilled Data Elucidate Model Biases.}\label{sec:interp}
\begin{figure}[t]
    \centering
    \includegraphics[width=1.0\linewidth]{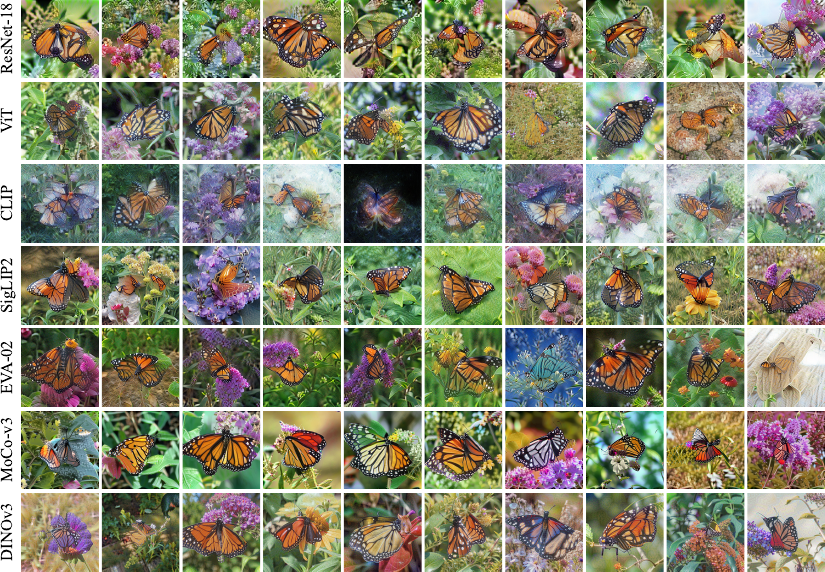}
    \caption{\looseness=-1 \textbf{Distilled images elucidate model representations.} 
    Each of the models shown here induces a different style of distilled images to best cover its own distribution. 
    CLIP~\cite{clip} is a notable outlier, possibly due to its strong text-entanglement and the contrastive loss not used in more recent vision-text models, such as SigLIP2~\cite{siglip2}.
    }
    \label{fig:models_compare}\vspace{-1.5em}
\end{figure}

One recent work (Linear Gradient Matching~\cite{cazenavette2025dataset}) focused specifically on distillation for pre-trained feature extractors. 
While this method highlighted the interactions between different models and their distilled data, it could not scale past 1 image per class due to massive resource requirements as a result of needing to jointly distill all classes together. 
We pick up where this work left off and distill 10 images per class with seven different backbones (ResNet-18~\cite{resnet}, ViT~\cite{vit}, CLIP~\cite{clip}, SigLIP2~\cite{siglip2}, EVA-02~\cite{fang2024eva}, MoCo-v3~\cite{mocov3}, and DINOv3~\cite{dinov3}). 
In Figure~\ref{fig:models_compare}, we see the monarch butterfly class distilled to 10 images per class for each of these models. 
As noted for 1 image per class in the linear gradient matching settings~\cite{cazenavette2025dataset}, each model yields images with its own particular bias despite using the same underlying distillation method. 
Most notably, we see that the CLIP images are quite murky and contain less-well-defined backgrounds than those of the other models, possibly due to the model's text entanglement. This is reflected quantitatively as well; when each set is evaluated in the other models' feature spaces (Appendix Table~\ref{tab:coverage_matrix}), CLIP-distilled data cover the least of the real distribution (on average) while DINOv3-distilled data cover the most.

\subsection{Drifting Distillation Generalizes to Other Modalities.}\label{sec:audiovideo}
\begin{figure}[t]
    \centering
    \begin{minipage}[b]{0.51\linewidth}
        \centering
        \includegraphics[width=\linewidth]{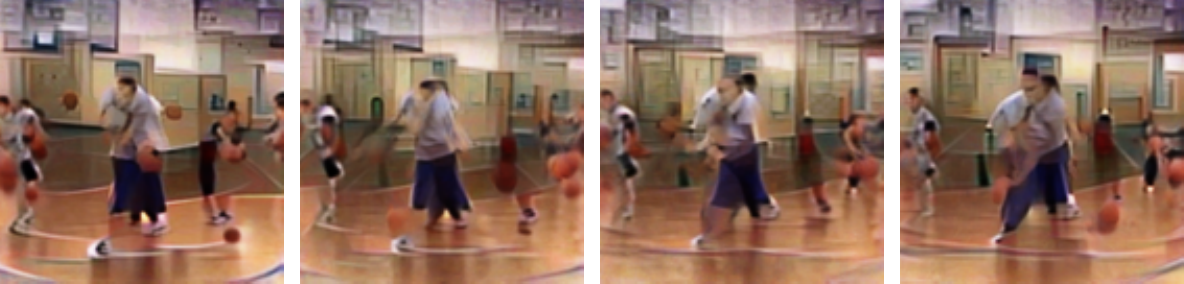}
        \captionof{figure}{\looseness=-1
        Four frames of a basketball dribbling video from HMDB51~\cite{kuehne2011hmdb} distilled via drifting with \mbox{VideoMAE-v2}~\cite{wang2023videomaev2} into a 3D Laplacian pyramid.}
        \label{fig:vid_grid}
    \end{minipage}
    \hfill
    \begin{minipage}[b]{0.47\linewidth}
    \small
        \centering
        \begin{tabular}{lcccc}
\toprule
 & \multicolumn{2}{c}{Audio} & \multicolumn{2}{c}{Video} \\
\cmidrule(lr){2-3}\cmidrule(lr){4-5}
Method & Linear & $k$-NN & Linear & $k$-NN \\
\midrule
  Herding & 45.6 & 39.9 & 66.6 & 60.4 \\
  Ours & \textbf{51.4} & \textbf{52.6} & \textbf{70.3} & \textbf{65.8} \\
\bottomrule
\end{tabular}

        \captionof{table}{\looseness=-1 Our drifting method generalizes to both \mbox{audio} and video, outperforming the coreset baseline.}
        \label{tab:audio-video}
    \end{minipage}
    \vspace{-0.5em}
\end{figure}

We find that our drifting method and pyramid parameterization generalize to both \textbf{audio} and \textbf{video} data. 
For audio, we distill the FMA~\cite{fma_dataset} small dataset of 8 music genres using the \mbox{MusicFM-FMA~\cite{won2023musicfm}} feature extraction model. 
For video, we distill the HMDB51~\cite{kuehne2011hmdb} human action recognition dataset of 51 action categories using the VideoMAE-v2~\cite{wang2023videomaev2} feature extraction model. 
 We distill 5 samples per class and evaluate with a linear probe and cosine 1-NN classification on mean-pooled features.
For both modalities, we compare to and outperform the herding~\cite{welling2009herding} coreset selection baseline (Table~\ref{tab:audio-video}). Figure~\ref{fig:vid_grid} shows frames from one of our distilled ``dribbling'' videos.

\subsection{An Image is Worth a Thousand Paintings.}\label{sec:art}

\begin{figure}[t]
    \centering
    \setlength{\tabcolsep}{1pt}
    \renewcommand{\arraystretch}{0}
    \begin{tabular}{cc}
        \includegraphics[width=0.49\linewidth]{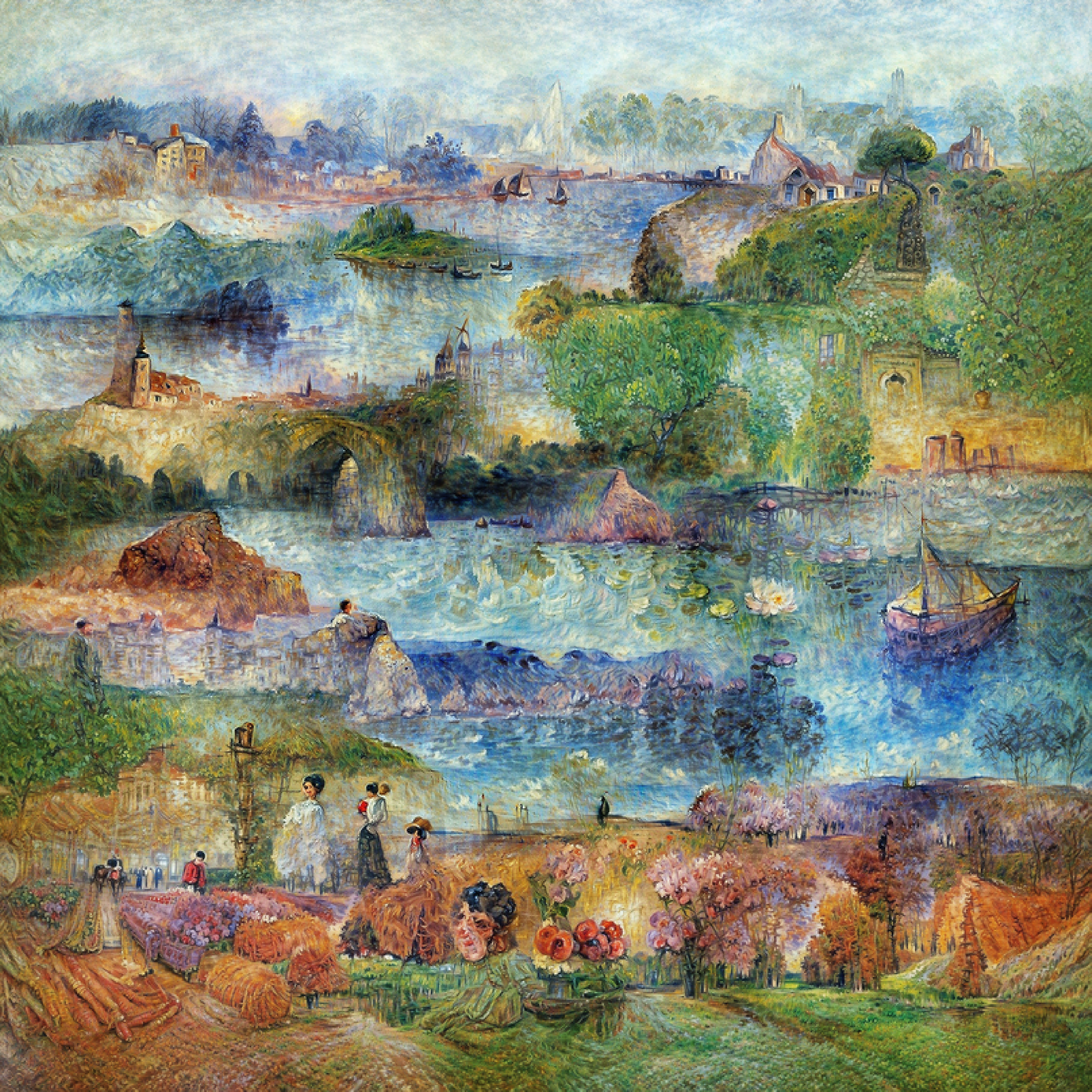}&\includegraphics[width=0.49\linewidth]{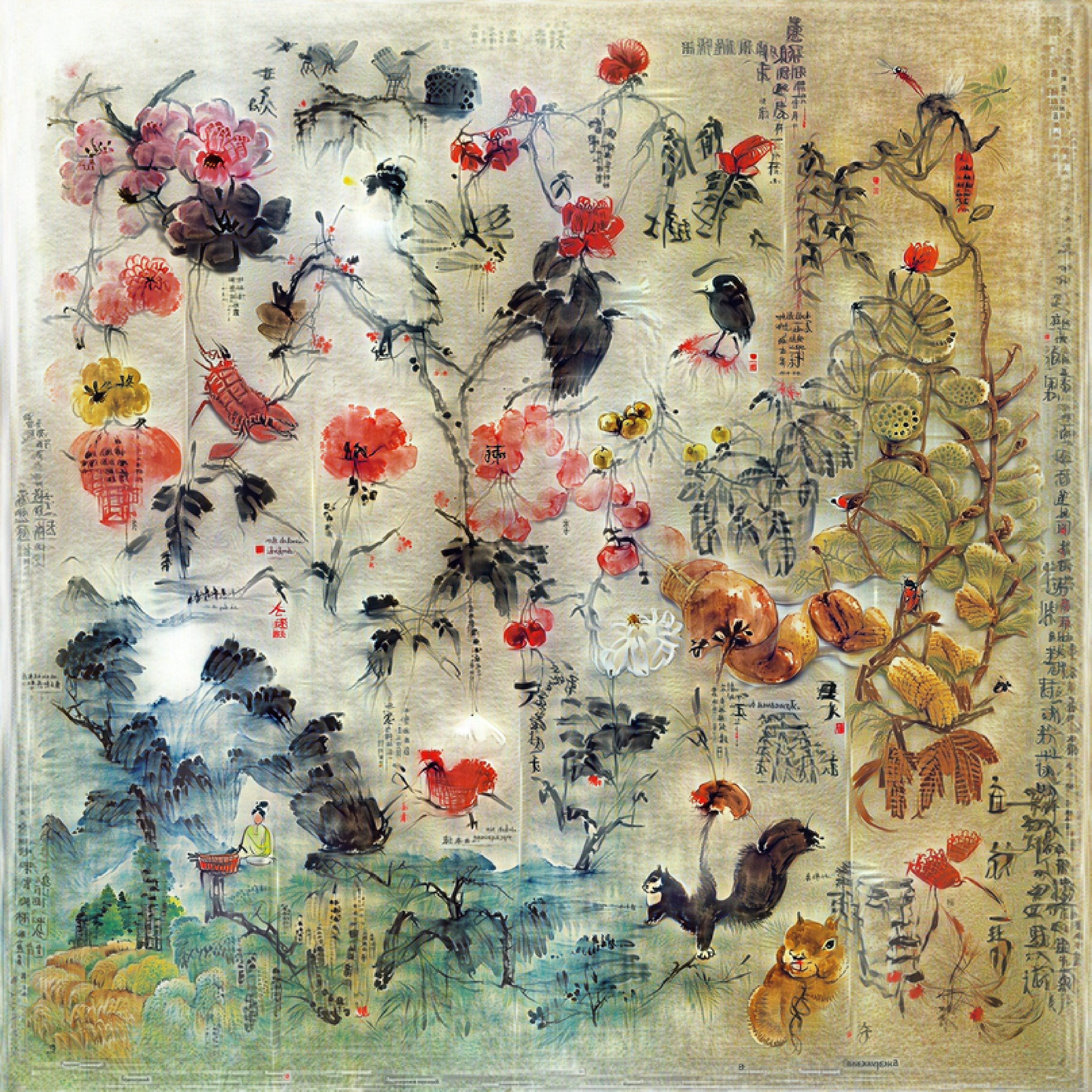}
    \end{tabular}
    \caption{
    \looseness=-1
    \textbf{Distilling Canvases.} We can distill a collection of images into a single large canvas via drifting by using a large number of random crops at each step. The above $2048\times2048$ pixel images were synthesized by drifting SigLIP2~\cite{siglip2} features to match the distribution of Claude Monet's (\textbf{left}) and Qi Baishi's (\textbf{right}) paintings.}
    \label{fig:art}
\end{figure}

While dataset distillation's artistic value has been briefly explored in the past~\cite{cazenavette2022wearable}, we find that our drifting method excels at distilling the essence of an art style into a large canvas~\cite{podd}. 
In Figure~\ref{fig:art}, we see the paintings of Claude Monet and Qi Baishi distilled into single images using \mbox{SigLIP2}~\cite{siglip2}. 
Remarkably, each distilled image clearly reflects its painter's individual style and contains common elements found throughout their work (e.g., Monet's water lilies and Qi's squirrels).

\section{Conclusion}
In this work, we introduced a new method of \textbf{Dataset Distillation via Drifting}, designed to explicitly learn a synthetic set of images that \textit{best covers the target distribution}. 
In addition to best approximating the real data distribution when compared to the baseline methods, our distilled images also serve as superior training data for classification models from scratch.

Furthermore, our drifting method offers insights into model interpretability, generalizes to audio and video, and can be used to synthesize artwork. 
We hope our work inspires further dataset distillation research, especially in the direction of distribution coverage.

\clearpage
\paragraph{Acknowledgments.} 
\looseness=-1
This work was supported by the National Science Foundation under Grant No. 2211259, by the Intelligence Advanced Research Projects Activity (IARPA) via Department of Interior/Interior Business Center (DOI/IBC) under 140D0423C0075, by the Amazon Science Hub, by the MIT-Google Program for Computing Innovation, by AMD via the MIT AI Hardware Program, by an ONR MURI grant 037564-00001, and by a 2025 MIT Office of Research Computing and Data Seed Grant. 
Giannis Daras is supported by a Jane Street Research Collaboration award, a TPU Research Cloud grant, and Lambda's Research Grant Program. We thank Mingyang Deng for sharing an early version of his drifting code with us and other members of the MIT community for providing valuable feedback on this work.

{\small
\bibliographystyle{abbrv}
\bibliography{neurips_2026}

\begin{thebibliography}{10}

\bibitem{pytorch2}
J.~Ansel, E.~Yang, H.~He, N.~Gimelshein, A.~Jain, M.~Voznesensky, B.~Bao, P.~Bell, D.~Berard, E.~Burovski, G.~Chauhan, A.~Chourdia, W.~Constable, A.~Desmaison, Z.~DeVito, E.~Ellison, W.~Feng, J.~Gong, M.~Gschwind, B.~Hirsh, S.~Huang, K.~Kalambarkar, L.~Kirsch, M.~Lazos, M.~Lezcano, Y.~Liang, J.~Liang, Y.~Lu, C.~K. Luk, B.~Maher, Y.~Pan, C.~Puhrsch, M.~Reso, M.~Saroufim, M.~Y. Siraichi, H.~Suk, S.~Zhang, M.~Suo, P.~Tillet, X.~Zhao, E.~Wang, K.~Zhou, R.~Zou, X.~Wang, A.~Mathews, W.~Wen, G.~Chanan, P.~Wu, and S.~Chintala.
\newblock {PyTorch} 2: Faster machine learning through dynamic {Python} bytecode transformation and graph compilation.
\newblock In {\em ASPLOS}, 2024.

\bibitem{borsos2020coresets}
Z.~Borsos, M.~Mutny, and A.~Krause.
\newblock Coresets via bilevel optimization for continual learning and streaming.
\newblock {\em NeurIPS}, 2020.

\bibitem{cazenavette2025dataset}
G.~Cazenavette, A.~Torralba, and V.~Sitzmann.
\newblock Dataset distillation for pre-trained self-supervised vision models.
\newblock {\em NeurIPS}, 2025.

\bibitem{mtt}
G.~Cazenavette, T.~Wang, A.~Torralba, A.~A. Efros, and J.-Y. Zhu.
\newblock Dataset distillation by matching training trajectories.
\newblock In {\em CVPR}, 2022.

\bibitem{cazenavette2022wearable}
G.~Cazenavette, T.~Wang, A.~Torralba, A.~A. Efros, and J.-Y. Zhu.
\newblock Wearable {ImageNet}: Synthesizing tileable textures via dataset distillation.
\newblock In {\em CVPR}, 2022.

\bibitem{glad}
G.~Cazenavette, T.~Wang, A.~Torralba, A.~A. Efros, and J.-Y. Zhu.
\newblock Generalizing dataset distillation via deep generative prior.
\newblock In {\em CVPR}, 2023.

\bibitem{mocov3}
X.~Chen, S.~Xie, and K.~He.
\newblock An empirical study of training self-supervised vision transformers.
\newblock In {\em ICCV}, 2021.

\bibitem{fadrm}
J.~Cui, X.~Bi, Y.~Luo, X.~Zhao, J.~Liu, and Z.~Shen.
\newblock {FADRM}: Fast and accurate data residual matching for dataset distillation.
\newblock In {\em NeurIPS}, 2025.

\bibitem{cvdd}
J.~Cui, Z.~Li, X.~Ma, X.~Bi, Y.~Luo, and Z.~Shen.
\newblock Dataset distillation via committee voting.
\newblock {\em arXiv}, 2025.

\bibitem{tesla}
J.~Cui, R.~Wang, S.~Si, and C.-J. Hsieh.
\newblock Scaling up dataset distillation to {ImageNet-1K} with constant memory.
\newblock In {\em ICML}, 2023.

\bibitem{fma_dataset}
M.~Defferrard, K.~Benzi, P.~Vandergheynst, and X.~Bresson.
\newblock {FMA}: A dataset for music analysis.
\newblock In {\em ISMIR}, 2017.

\bibitem{imagenet1k}
J.~Deng, W.~Dong, R.~Socher, L.-J. Li, K.~Li, and L.~Fei-Fei.
\newblock {ImageNet}: A large-scale hierarchical image database.
\newblock In {\em CVPR}, 2009.

\bibitem{drifting}
M.~Deng, H.~Li, T.~Li, Y.~Du, and K.~He.
\newblock Generative modeling via drifting.
\newblock In {\em NeurIPS}, 2026.

\bibitem{dey2026rethinking}
P.~Dey, A.~Sahdev, S.~Bhati, K.~R. Mopuri, and R.~V. Babu.
\newblock Rethinking dataset distillation: Hard truths about soft labels.
\newblock In {\em CVPR}, 2026.

\bibitem{dong2022privacy}
T.~Dong, B.~Zhao, and L.~Liu.
\newblock Privacy for free: How does dataset condensation help privacy?
\newblock In {\em ICML}, 2022.

\bibitem{vit}
A.~Dosovitskiy, L.~Beyer, A.~Kolesnikov, D.~Weissenborn, X.~Zhai, T.~Unterthiner, M.~Dehghani, M.~Minderer, G.~Heigold, S.~Gelly, et~al.
\newblock An image is worth 16x16 words: Transformers for image recognition at scale.
\newblock In {\em ICLR}, 2020.

\bibitem{dwa}
J.~Du, X.~Zhang, J.~Hu, W.~Huang, and J.~T. Zhou.
\newblock Diversity-driven synthesis: Enhancing dataset distillation through directed weight adjustment.
\newblock {\em NeurIPS}, 2024.

\bibitem{fang2024eva}
Y.~Fang, Q.~Sun, X.~Wang, T.~Huang, X.~Wang, and Y.~Cao.
\newblock {EVA-02}: A visual representation for neon genesis.
\newblock {\em Image and Vision Computing}, 2024.

\bibitem{imagenette}
Fastai.
\newblock Fastai/imagenette: A smaller subset of 10 easily classified classes from imagenet, and a little more french.

\bibitem{feldman2011unified}
D.~Feldman and M.~Langberg.
\newblock A unified framework for approximating and clustering data.
\newblock In {\em STOC}, 2011.

\bibitem{fort2025direct}
S.~Fort and J.~Whitaker.
\newblock Direct ascent synthesis: Revealing hidden generative capabilities in discriminative models.
\newblock {\em arXiv}, 2025.

\bibitem{goodfellow2020generative}
I.~Goodfellow, J.~Pouget-Abadie, M.~Mirza, B.~Xu, D.~Warde-Farley, S.~Ozair, A.~Courville, and Y.~Bengio.
\newblock Generative adversarial networks.
\newblock {\em Communications of the ACM}, 63(11), 2020.

\bibitem{minimax}
J.~Gu, S.~Vahidian, V.~Kungurtsev, H.~Wang, W.~Jiang, Y.~You, and Y.~Chen.
\newblock Efficient dataset distillation via minimax diffusion.
\newblock In {\em CVPR}, 2024.

\bibitem{datm}
Z.~Guo, K.~Wang, G.~Cazenavette, H.~Li, K.~Zhang, and Y.~You.
\newblock Towards lossless dataset distillation via difficulty-aligned trajectory matching.
\newblock In {\em ICLR}, 2024.

\bibitem{har2004coresets}
S.~Har-Peled and S.~Mazumdar.
\newblock On coresets for k-means and k-median clustering.
\newblock In {\em STOC}, 2004.

\bibitem{resnet}
K.~He, X.~Zhang, S.~Ren, and J.~Sun.
\newblock Deep residual learning for image recognition.
\newblock In {\em CVPR}, 2016.

\bibitem{hinton2015distilling}
G.~Hinton, O.~Vinyals, and J.~Dean.
\newblock Distilling the knowledge in a neural network.
\newblock In {\em NeurIPS Workshop}, 2015.

\bibitem{ho2020denoising}
J.~Ho, A.~Jain, and P.~Abbeel.
\newblock Denoising diffusion probabilistic models.
\newblock {\em NeurIPS}, 2020.

\bibitem{hoogeboom2023simple}
E.~Hoogeboom, J.~Heek, and T.~Salimans.
\newblock simple diffusion: End-to-end diffusion for high resolution images.
\newblock In {\em ICML}, 2023.

\bibitem{huggingface2016}
{Hugging Face}.
\newblock {Hugging Face: The AI community building the future}.
\newblock \url{https://huggingface.co}, 2016.

\bibitem{kingma2014adam}
D.~Kingma and J.~Ba.
\newblock Adam: A method for stochastic optimization.
\newblock In {\em ICLR}, 2015.

\bibitem{kuehne2011hmdb}
H.~Kuehne, H.~Jhuang, E.~Garrote, T.~Poggio, and T.~Serre.
\newblock {HMDB}: A large video database for human motion recognition.
\newblock In {\em ICCV}, 2011.

\bibitem{reta}
M.~Li, H.~Gou, Y.~Ma, R.~Wang, K.~Qin, and T.~He.
\newblock Fixed anchors are not enough: Dynamic retrieval and persistent homology for dataset distillation.
\newblock In {\em CVPR}, 2026.

\bibitem{lipman2022flow}
Y.~Lipman, R.~T. Chen, H.~Ben-Hamu, M.~Nickel, and M.~Le.
\newblock Flow matching for generative modeling.
\newblock In {\em ICLR}, 2023.

\bibitem{wmdd}
H.~Liu, Y.~Li, T.~Xing, P.~Wang, V.~Dalal, L.~Li, J.~He, and H.~Wang.
\newblock Dataset distillation via the {W}asserstein metric.
\newblock In {\em ICCV}, 2025.

\bibitem{cudd}
Z.~Ma, A.~Cao, F.~Yang, Y.~Gong, and X.~Wei.
\newblock Curriculum dataset distillation.
\newblock {\em TIP}, 2025.

\bibitem{mirzasoleiman2020coresets}
B.~Mirzasoleiman, K.~Cao, and J.~Leskovec.
\newblock Coresets for robust training of deep neural networks against noisy labels.
\newblock {\em NeurIPS}, 2020.

\bibitem{dire}
S.~Mohanty, A.~Reddy, and K.~R. Mopuri.
\newblock {DiRe}: Diversity-promoting regularization for dataset condensation.
\newblock In {\em WACV}, 2026.

\bibitem{deepdream}
A.~Mordvintsev, C.~Olah, and M.~Tyka.
\newblock {DeepDream} -- a code example for visualizing neural networks.
\newblock Google Research Blog, jul 2015.

\bibitem{prism}
B.~B. Moser, S.~Sarode, F.~Raue, S.~Frolov, K.~Adamkiewicz, A.~Shanbhag, J.~Folz, T.~C. Nauen, and A.~Dengel.
\newblock {PRISM}: Diversifying dataset distillation by decoupling architectural priors.
\newblock {\em TMLR}, 2026.

\bibitem{naeem2020reliable}
M.~F. Naeem, S.~J. Oh, Y.~Uh, Y.~Choi, and J.~Yoo.
\newblock Reliable fidelity and diversity metrics for generative models.
\newblock In {\em ICML}, 2020.

\bibitem{olah2017feature}
C.~Olah, A.~Mordvintsev, and L.~Schubert.
\newblock Feature visualization.
\newblock {\em Distill}, 2017.
\newblock https://distill.pub/2017/feature-visualization.

\bibitem{pytorch}
A.~Paszke, S.~Gross, F.~Massa, A.~Lerer, J.~Bradbury, G.~Chanan, T.~Killeen, Z.~Lin, N.~Gimelshein, L.~Antiga, A.~Desmaison, A.~Kopf, E.~Yang, Z.~DeVito, M.~Raison, A.~Tejani, S.~Chilamkurthy, B.~Steiner, L.~Fang, J.~Bai, and S.~Chintala.
\newblock {PyTorch}: An imperative style, high-performance deep learning library.
\newblock In {\em NeurIPS}. 2019.

\bibitem{paul2021deep}
M.~Paul, S.~Ganguli, and G.~K. Dziugaite.
\newblock Deep learning on a data diet: Finding important examples early in training.
\newblock {\em NeurIPS}, 2021.

\bibitem{clip}
A.~Radford, J.~W. Kim, C.~Hallacy, A.~Ramesh, G.~Goh, S.~Agarwal, G.~Sastry, A.~Askell, P.~Mishkin, J.~Clark, et~al.
\newblock Learning transferable visual models from natural language supervision.
\newblock In {\em ICML}, 2021.

\bibitem{rebuffi2017icarl}
S.-A. Rebuffi, A.~Kolesnikov, G.~Sperl, and C.~H. Lampert.
\newblock {iCaRL}: Incremental classifier and representation learning.
\newblock In {\em CVPR}, 2017.

\bibitem{manifoldgd}
A.~Roy, W.-Y.~A. Lee, R.~Chakraborty, and V.~S. Lokhande.
\newblock {ManifoldGD}: Training-free hierarchical manifold guidance for diffusion-based dataset distillation.
\newblock In {\em CVPR}, 2026.

\bibitem{sener2017active}
O.~Sener and S.~Savarese.
\newblock Active learning for convolutional neural networks: A core-set approach.
\newblock In {\em ICLR}, 2018.

\bibitem{gvbsm}
S.~Shao, Z.~Yin, M.~Zhou, X.~Zhang, and Z.~Shen.
\newblock Generalized large-scale data condensation via various backbone and statistical matching.
\newblock In {\em CVPR}, 2024.

\bibitem{edc}
S.~Shao, Z.~Zhou, H.~Chen, and Z.~Shen.
\newblock Elucidating the design space of dataset condensation.
\newblock {\em NeurIPS}, 2024.

\bibitem{delt}
Z.~Shen, A.~Sherif, Z.~Yin, and S.~Shao.
\newblock {DELT}: A simple diversity-driven {EarlyLate} training for dataset distillation.
\newblock In {\em CVPR}, 2025.

\bibitem{podd}
A.~Shul, E.~Horwitz, and Y.~Hoshen.
\newblock Distilling datasets into less than one image.
\newblock {\em TMLR}, 2025.

\bibitem{dinov3}
O.~Sim{\'e}oni, H.~V. Vo, M.~Seitzer, F.~Baldassarre, M.~Oquab, C.~Jose, V.~Khalidov, M.~Szafraniec, S.~Yi, M.~Ramamonjisoa, F.~Massa, D.~Haziza, L.~Wehrstedt, J.~Wang, T.~Darcet, T.~Moutakanni, L.~Sentana, C.~Roberts, A.~Vedaldi, J.~Tolan, J.~Brandt, C.~Couprie, J.~Mairal, H.~J{\'e}gou, P.~Labatut, and P.~Bojanowski.
\newblock {DINO}v3.
\newblock {\em arXiv}, 2025.

\bibitem{sohl2015deep}
J.~Sohl-Dickstein, E.~Weiss, N.~Maheswaranathan, and S.~Ganguli.
\newblock Deep unsupervised learning using nonequilibrium thermodynamics.
\newblock In {\em ICML}, 2015.

\bibitem{d4m}
D.~Su, J.~Hou, W.~Gao, Y.~Tian, and B.~Tang.
\newblock {D$^4$M}: Dataset distillation via disentangled diffusion model.
\newblock In {\em CVPR}, 2024.

\bibitem{rded}
P.~Sun, B.~Shi, D.~Yu, and T.~Lin.
\newblock On the diversity and realism of distilled dataset: An efficient dataset distillation paradigm.
\newblock In {\em CVPR}, 2024.

\bibitem{toneva2018empirical}
M.~Toneva, A.~Sordoni, R.~T.~d. Combes, A.~Trischler, Y.~Bengio, and G.~J. Gordon.
\newblock An empirical study of example forgetting during deep neural network learning.
\newblock In {\em ICLR}, 2019.

\bibitem{torchvision2016}
{\relax TorchVision maintainers and contributors}.
\newblock {TorchVision}: {PyTorch}'s computer vision library.
\newblock \url{https://github.com/pytorch/vision}, 2016.

\bibitem{nrr}
M.-T. Tran, T.~Le, X.-M. Le, T.-T. Do, and D.~Phung.
\newblock Enhancing dataset distillation via non-critical region refinement.
\newblock In {\em CVPR}, 2025.

\bibitem{siglip2}
M.~Tschannen, A.~Gritsenko, X.~Wang, M.~F. Naeem, I.~Alabdulmohsin, N.~Parthasarathy, T.~Evans, L.~Beyer, Y.~Xia, B.~Mustafa, et~al.
\newblock {SigLIP} 2: Multilingual vision-language encoders with improved semantic understanding, localization, and dense features.
\newblock {\em arXiv}, 2025.

\bibitem{cao2}
H.~Wang, Z.~Zhao, J.~Wu, Y.~Shang, G.~Liu, and Y.~Yan.
\newblock {CaO$_2$}: Rectifying inconsistencies in diffusion-based dataset distillation.
\newblock In {\em ICCV}, 2025.

\bibitem{wang2022cafe}
K.~Wang, B.~Zhao, X.~Peng, Z.~Zhu, S.~Yang, S.~Wang, G.~Huang, H.~Bilen, X.~Wang, and Y.~You.
\newblock {CAFE}: Learning to condense dataset by aligning features.
\newblock {\em CVPR}, 2022.

\bibitem{wang2023videomaev2}
L.~Wang, B.~Huang, Z.~Zhao, Z.~Tong, Y.~He, Y.~Wang, Y.~Wang, and Y.~Qiao.
\newblock {VideoMAE V2}: Scaling video masked autoencoders with dual masking.
\newblock In {\em CVPR}, 2023.

\bibitem{infoutil}
S.~Wang, Y.~Yang, G.~Chen, P.~Li, K.~Li, Y.~Zhou, Z.~Chen, and L.~Zhang.
\newblock Grounding and enhancing informativeness and utility in dataset distillation.
\newblock In {\em ICLR}, 2026.

\bibitem{dd}
T.~Wang, J.-Y. Zhu, A.~Torralba, and A.~A. Efros.
\newblock Dataset distillation.
\newblock {\em arXiv}, 2018.

\bibitem{welling2009herding}
M.~Welling.
\newblock Herding dynamical weights to learn.
\newblock In {\em ICML}, 2009.

\bibitem{rw2019timm}
R.~Wightman.
\newblock {PyTorch} image models.
\newblock \url{https://huggingface.co/timm}, 2019.

\bibitem{won2023musicfm}
M.~Won, Y.-N. Hung, and D.~Le.
\newblock A foundation model for music informatics.
\newblock In {\em ICASSP}, 2024.

\bibitem{yang2024mind}
S.~Yang, Z.~Cao, S.~Guo, R.~Zhang, P.~Luo, S.~Zhang, and L.~Nie.
\newblock Mind the boundary: Coreset selection via reconstructing the decision boundary.
\newblock In {\em ICML}, 2024.

\bibitem{deepinversion}
H.~Yin, P.~Molchanov, J.~M. Alvarez, Z.~Li, A.~Mallya, D.~Hoiem, N.~K. Jha, and J.~Kautz.
\newblock Dreaming to distill: Data-free knowledge transfer via {DeepInversion}.
\newblock In {\em CVPR}, 2020.

\bibitem{cda}
Z.~Yin and Z.~Shen.
\newblock Dataset distillation via curriculum data synthesis in large data era.
\newblock {\em TMLR}, 2024.

\bibitem{sre2l}
Z.~Yin, E.~Xing, and Z.~Shen.
\newblock Squeeze, recover and relabel: Dataset condensation at {ImageNet} scale from a new perspective.
\newblock {\em NeurIPS}, 2023.

\bibitem{ufc}
X.~Zhang, J.~Du, P.~Liu, and J.~T. Zhou.
\newblock Breaking class barriers: Efficient dataset distillation via inter-class feature compensator.
\newblock In {\em ICLR}, 2025.

\bibitem{dsa}
B.~Zhao and H.~Bilen.
\newblock Dataset condensation with differentiable {S}iamese augmentation.
\newblock In {\em ICML}, 2021.

\bibitem{dm}
B.~Zhao and H.~Bilen.
\newblock Dataset condensation with distribution matching.
\newblock {\em WACV}, 2023.

\bibitem{dc}
B.~Zhao, K.~R. Mopuri, and H.~Bilen.
\newblock Dataset condensation with gradient matching.
\newblock In {\em ICLR}, 2020.

\bibitem{d3hr}
L.~Zhao, Y.~Wu, X.~Jiang, J.~Gu, Y.~Wang, X.~Xu, P.~Zhao, and X.~Lin.
\newblock Taming diffusion for dataset distillation with high representativeness.
\newblock In {\em ICML}, 2025.

\bibitem{scdd}
M.~Zhou, Z.~Yin, S.~Shao, and Z.~Shen.
\newblock Self-supervised dataset distillation: A good compression is all you need.
\newblock {\em arXiv}, 2024.

\bibitem{eupe}
C.~Zhu, S.~Suri, C.~Jose, M.~Oquab, M.~Szafraniec, W.~Wen, Y.~Xiong, P.~Labatut, P.~Bojanowski, R.~Krishnamoorthi, and V.~Chandra.
\newblock Efficient universal perception encoder.
\newblock {\em arXiv}, 2026.

\end{thebibliography}
}
\clearpage
\appendix
\counterwithin{figure}{section}
\counterwithin{table}{section}

\counterwithin{figure}{section}
\counterwithin{figure}{section}
\counterwithin{table}{section}
\section*{Appendix}
\suppressfloats[t]  %
\section{Hyperparameter Sweeps}\label{sec:ablations}
\setlength{\intextsep}{3pt plus 1fill minus 2pt}
Each study below changes a single setting of our default configuration (DINOv3 backbone, noise initialization, ImageNet-1k, IPC 10) and reports the same from-scratch and feature-space metrics as Tables~\ref{tab:from_scratch} and~\ref{tab:feature_space}; defaults are marked $\dagger$. Each table is paired with a figure showing all 10 distilled images of a single class.

\begin{figure}[H]
    \centering
    \includegraphics[width=0.95\linewidth]{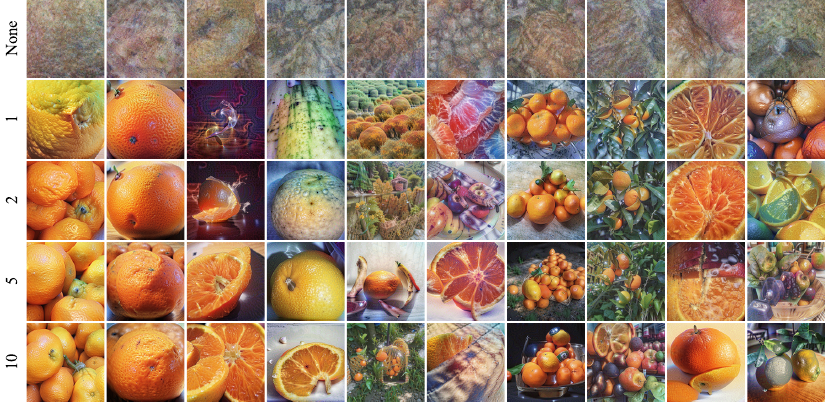}
    \caption{\looseness=-1 \textbf{Effects of augmentations.} 
    Here, we visualize the effect the number of augmentations per step has on the distilled data. 
    As we increase the number of augmentations, the images appear to (qualitatively) become richer in semantic information. 
    Images shown here were initialized with noise and distilled with DINOv3 to match ImageNet-1k's ``orange'' class.
    For quantitative results, please see Table~\ref{tab:abl_crops}.}
    \label{fig:compare_augs}
\end{figure}

\paragraph{Augmentations per step.} Drifting augmented views of each image is critical. Without augmentation, every metric collapses (e.g., 0.2\% hard-label accuracy and 1.0\% coverage). Additionally, every metric improves monotonically as more augmented views are drifted per step, with 10 views performing best throughout (Table~\ref{tab:abl_crops}, Figure~\ref{fig:compare_augs}).

\begin{table}[H]
\centering
\small
\setlength{\tabcolsep}{3pt}%
\begin{tabular}{l ccc cc cc cc}
\toprule
 & \multicolumn{3}{c}{ResNet-18 from scratch} & \multicolumn{2}{c}{Coverage} & \multicolumn{2}{c}{k-NN} & \multicolumn{2}{c}{Linear} \\
\cmidrule(lr){2-4}\cmidrule(lr){5-6}\cmidrule(lr){7-8}\cmidrule(lr){9-10}
 & Hard & Soft & KD & {\smaller SigLIP2} & {\smaller MoCo-v3} & {\smaller SigLIP2} & {\smaller MoCo-v3} & {\smaller SigLIP2} & {\smaller MoCo-v3} \\
\midrule
None & \phantom{0}0.2 \textcolor{gray}{$\pm$ 0.0} & \phantom{0}0.2 \textcolor{gray}{$\pm$ 0.0} & \phantom{0}1.9 \textcolor{gray}{$\pm$ 0.1} & \phantom{0}1.0 & \phantom{0}0.6 & \phantom{0}2.7 & \phantom{0}5.8 & \phantom{0}6.5 & \phantom{0}4.9 \\
1 & \phantom{0}7.5 \textcolor{gray}{$\pm$ 0.3} & 27.4 \textcolor{gray}{$\pm$ 0.3} & 39.3 \textcolor{gray}{$\pm$ 0.3} & 35.8 & 28.6 & 56.0 & 50.2 & 69.5 & 59.2 \\
2 & 15.5 \textcolor{gray}{$\pm$ 0.4} & 34.4 \textcolor{gray}{$\pm$ 0.2} & 42.4 \textcolor{gray}{$\pm$ 0.6} & 65.5 & 66.0 & 66.5 & 57.4 & 76.6 & 64.9 \\
5 & \underline{19.1} \textcolor{gray}{$\pm$ 0.4} & \underline{35.4} \textcolor{gray}{$\pm$ 0.3} & \underline{44.2} \textcolor{gray}{$\pm$ 0.4} & \underline{75.1} & \underline{79.9} & \underline{69.4} & \underline{60.3} & \underline{77.3} & \underline{66.1} \\
10$^\dagger$ & \textbf{21.3} \textcolor{gray}{$\pm$ 0.5} & \textbf{35.9} \textcolor{gray}{$\pm$ 0.1} & \textbf{45.0} \textcolor{gray}{$\pm$ 0.2} & \textbf{77.0} & \textbf{83.4} & \textbf{69.9} & \textbf{61.1} & \textbf{77.5} & \textbf{66.3} \\
\bottomrule
\end{tabular}

\vspace{0.5em}
\caption{\looseness=-1 \textbf{Augmented copies per step.} The number of augmented views of each distilled image used at every optimization step. \emph{None} is a single deterministic resized view. Performance monotonically increases with the number of augmentations, with diminishing returns as we approach 10.}
\label{tab:abl_crops}
\end{table}

\begin{figure}[H]
    \centering
    \includegraphics[width=0.95\linewidth]{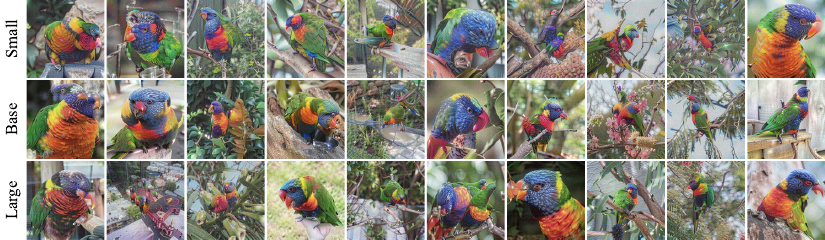}
    \caption{\looseness=-1 \textbf{Effect of backbone size.} All 10 distilled images of the ImageNet-1k ``lorikeet'' class using DINOv3 ViT-S, ViT-B (default), and ViT-L. For quantitative results, see Table~\ref{tab:abl_model_size}.}
    \label{fig:compare_sizes}
\end{figure}

\clearpage
\paragraph{Backbone size.} A larger backbone does not necessarily lead to better data. DINOv3 ViT-B is best or tied on every metric, while ViT-L matches it only in the knowledge distillation setting and ViT-S covers notably less of the distribution (Table~\ref{tab:abl_model_size}, Figure~\ref{fig:compare_sizes}). However, the ViT-L variant could possibly perform better with more hyperparameter tuning.

\begin{table}[H]
\centering
\small
\setlength{\tabcolsep}{3pt}%
\begin{tabular}{l ccc cc cc cc}
\toprule
 & \multicolumn{3}{c}{ResNet-18 from scratch} & \multicolumn{2}{c}{Coverage} & \multicolumn{2}{c}{k-NN} & \multicolumn{2}{c}{Linear} \\
\cmidrule(lr){2-4}\cmidrule(lr){5-6}\cmidrule(lr){7-8}\cmidrule(lr){9-10}
 & Hard & Soft & KD & {\smaller SigLIP2} & {\smaller MoCo-v3} & {\smaller SigLIP2} & {\smaller MoCo-v3} & {\smaller SigLIP2} & {\smaller MoCo-v3} \\
\midrule
Small & \underline{18.5} \textcolor{gray}{$\pm$ 0.5} & 31.9 \textcolor{gray}{$\pm$ 0.5} & 41.9 \textcolor{gray}{$\pm$ 0.5} & 61.8 & 77.4 & 67.6 & \underline{60.9} & \underline{77.2} & 65.3 \\
Base$^\dagger$ & \textbf{21.3} \textcolor{gray}{$\pm$ 0.5} & \textbf{35.9} \textcolor{gray}{$\pm$ 0.1} & \textbf{45.0} \textcolor{gray}{$\pm$ 0.2} & \textbf{77.0} & \textbf{83.4} & \textbf{69.9} & \textbf{61.1} & \textbf{77.5} & \textbf{66.3} \\
Large & \underline{18.3} \textcolor{gray}{$\pm$ 0.3} & \underline{34.0} \textcolor{gray}{$\pm$ 0.4} & \textbf{44.6} \textcolor{gray}{$\pm$ 0.4} & \underline{75.6} & \underline{78.1} & \underline{69.8} & 60.2 & 76.9 & \underline{65.6} \\
\bottomrule
\end{tabular}

\vspace{0.5em}
\caption{\looseness=-1 \textbf{Backbone model size.} Base size performs the best with our settings, but the larger model might improve with some tuning.}
\label{tab:abl_model_size}
\end{table}

\begin{figure}[H]
    \centering
    \includegraphics[width=0.95\linewidth]{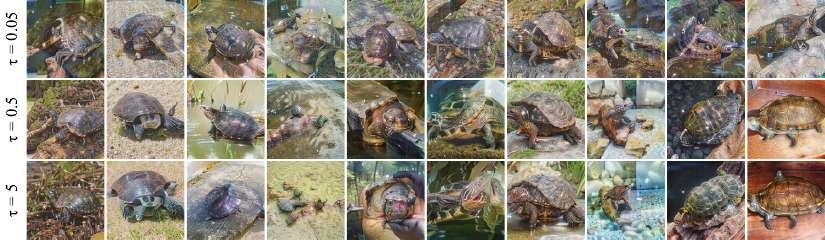}
    \caption{\looseness=-1 \textbf{Effect of kernel temperature.} All 10 distilled images of the ImageNet-1k ``terrapin'' class for three drifting kernel temperatures $\tau$ (default $\tau=0.5$). For quantitative results, see Table~\ref{tab:abl_temperature}.}
    \label{fig:compare_temps}
\end{figure}

\paragraph{Kernel temperature.} A small temperature ($\tau=0.05$) slightly improves $k$-NN accuracy but substantially lowers soft-label accuracy and coverage, while a large temperature ($\tau=5$) matches the default in the soft-label and knowledge distillation settings but lowers hard-label accuracy and coverage. Our default ($\tau=0.5$) gives the best balance (Table~\ref{tab:abl_temperature}, Figure~\ref{fig:compare_temps}).

\begin{table}[H]
\centering
\small
\setlength{\tabcolsep}{3pt}%
\begin{tabular}{l ccc cc cc cc}
\toprule
 & \multicolumn{3}{c}{ResNet-18 from scratch} & \multicolumn{2}{c}{Coverage} & \multicolumn{2}{c}{k-NN} & \multicolumn{2}{c}{Linear} \\
\cmidrule(lr){2-4}\cmidrule(lr){5-6}\cmidrule(lr){7-8}\cmidrule(lr){9-10}
 & Hard & Soft & KD & {\smaller SigLIP2} & {\smaller MoCo-v3} & {\smaller SigLIP2} & {\smaller MoCo-v3} & {\smaller SigLIP2} & {\smaller MoCo-v3} \\
\midrule
$\tau = 0.05$ & \underline{17.9} \textcolor{gray}{$\pm$ 0.2} & 26.4 \textcolor{gray}{$\pm$ 0.1} & 39.6 \textcolor{gray}{$\pm$ 0.2} & \underline{71.6} & 66.6 & \textbf{72.7} & \textbf{62.7} & \underline{76.9} & 64.4 \\
$\tau = 0.5^\dagger$ & \textbf{21.3} \textcolor{gray}{$\pm$ 0.5} & \textbf{35.9} \textcolor{gray}{$\pm$ 0.1} & \textbf{45.0} \textcolor{gray}{$\pm$ 0.2} & \textbf{77.0} & \textbf{83.4} & \underline{69.9} & \underline{61.1} & \textbf{77.5} & \textbf{66.3} \\
$\tau = 5$ & \underline{17.7} \textcolor{gray}{$\pm$ 0.1} & \textbf{35.9} \textcolor{gray}{$\pm$ 0.4} & \textbf{45.1} \textcolor{gray}{$\pm$ 0.5} & 64.7 & \underline{70.7} & 66.2 & 57.8 & 76.1 & \underline{65.0} \\
\bottomrule
\end{tabular}

\vspace{0.5em}
\caption{\looseness=-1\textbf{Drifting distance kernel temperature.} The method remains decently robust to a higher temperature, but a lower one significantly hurts performance.}
\label{tab:abl_temperature}
\end{table}

\begin{figure}[H]
    \centering
    \includegraphics[width=0.95\linewidth]{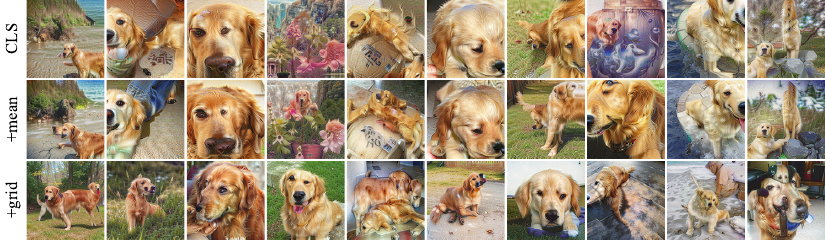}
    \caption{\looseness=-1 \textbf{Effect of the drifted features.} All 10 distilled images of the ImageNet-1k ``golden retriever'' class when the drifting loss is applied to the CLS token only, the CLS token plus the mean patch token, or the CLS token plus every patch token (default). For quantitative results, see Table~\ref{tab:abl_drifted_features}.}
    \label{fig:drifted_features}
\end{figure}

\clearpage
\paragraph{Drifted features.} Drifting the CLS token alone already yields usable data, but adding the mean patch token and, even more so, \textit{every} patch token steadily improves from-scratch accuracy (11.8\% $\rightarrow$ 15.0\% $\rightarrow$ 21.3\% hard-label) and coverage, while $k$-NN and linear accuracy remain nearly unchanged (Table~\ref{tab:abl_drifted_features}, Figure~\ref{fig:drifted_features}).

\begin{table}[H]
\centering
\small
\setlength{\tabcolsep}{3pt}%
\begin{tabular}{l ccc cc cc cc}
\toprule
 & \multicolumn{3}{c}{ResNet-18 from scratch} & \multicolumn{2}{c}{Coverage} & \multicolumn{2}{c}{k-NN} & \multicolumn{2}{c}{Linear} \\
\cmidrule(lr){2-4}\cmidrule(lr){5-6}\cmidrule(lr){7-8}\cmidrule(lr){9-10}
 & Hard & Soft & KD & {\smaller SigLIP2} & {\smaller MoCo-v3} & {\smaller SigLIP2} & {\smaller MoCo-v3} & {\smaller SigLIP2} & {\smaller MoCo-v3} \\
\midrule
{\smaller CLS} & 11.8 \textcolor{gray}{$\pm$ 0.2} & 26.4 \textcolor{gray}{$\pm$ 0.0} & 41.3 \textcolor{gray}{$\pm$ 0.3} & 70.2 & 72.5 & \underline{69.4} & \textbf{61.8} & 76.7 & 66.0 \\
{\smaller $+$mean} & \underline{15.0} \textcolor{gray}{$\pm$ 0.3} & \underline{31.0} \textcolor{gray}{$\pm$ 0.3} & \underline{43.6} \textcolor{gray}{$\pm$ 0.6} & \underline{74.2} & \underline{80.0} & 69.3 & \underline{61.3} & \underline{77.4} & \textbf{66.6} \\
{\smaller $+$grid$^\dagger$} & \textbf{21.3} \textcolor{gray}{$\pm$ 0.5} & \textbf{35.9} \textcolor{gray}{$\pm$ 0.1} & \textbf{45.0} \textcolor{gray}{$\pm$ 0.2} & \textbf{77.0} & \textbf{83.4} & \textbf{69.9} & 61.1 & \textbf{77.5} & \underline{66.3} \\
\bottomrule
\end{tabular}

\vspace{0.5em}
\caption{\looseness=-1 \textbf{Drifted features.} The DINOv3 tokens to which the drifting loss is applied: the CLS token only, the CLS token plus the mean patch token ($+$mean), or the CLS token plus every patch token ($+$grid). Drifting the CLS token with the full grid performs best.}
\label{tab:abl_drifted_features}
\end{table}

\begin{figure}[H]
    \centering
    \includegraphics[width=0.95\linewidth]{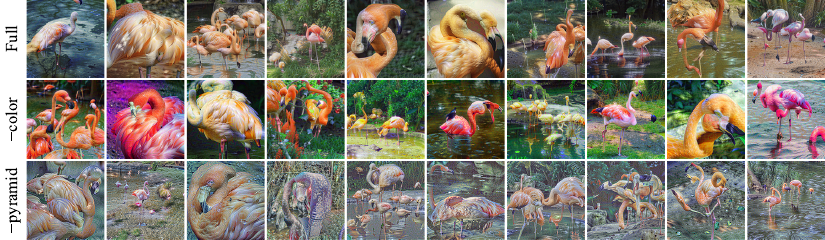}
    \caption{\looseness=-1 \textbf{Effect of the parameterization.} All 10 distilled images of the ImageNet-1k ``flamingo'' class with our full parameterization, without color decorrelation ($-$color), and with raw pixels instead of a Laplacian pyramid ($-$pyramid). For quantitative results, see Table~\ref{tab:abl_parameterization}.}
    \label{fig:parameterization}
\end{figure}

\paragraph{Parameterization.} Both components of our parameterization matter. Removing color decorrelation lowers hard-label accuracy from 21.3\% to 16.5\% and SigLIP2 coverage from 77.0\% to 62.0\%, and replacing the Laplacian pyramid with raw pixels collapses from-scratch accuracy (3.3\% hard-label) and coverage (41.9\%). Notably, $k$-NN and linear accuracy barely change in either case, again showing that these metrics are much less sensitive to the quality of the distilled data than from-scratch training or coverage (Table~\ref{tab:abl_parameterization}, Figure~\ref{fig:parameterization}).

\begin{table}[H]
\centering
\small
\setlength{\tabcolsep}{3pt}%
\begin{tabular}{l ccc cc cc cc}
\toprule
 & \multicolumn{3}{c}{ResNet-18 from scratch} & \multicolumn{2}{c}{Coverage} & \multicolumn{2}{c}{k-NN} & \multicolumn{2}{c}{Linear} \\
\cmidrule(lr){2-4}\cmidrule(lr){5-6}\cmidrule(lr){7-8}\cmidrule(lr){9-10}
 & Hard & Soft & KD & {\smaller SigLIP2} & {\smaller MoCo-v3} & {\smaller SigLIP2} & {\smaller MoCo-v3} & {\smaller SigLIP2} & {\smaller MoCo-v3} \\
\midrule
{\smaller Full$^\dagger$} & \textbf{21.3} \textcolor{gray}{$\pm$ 0.5} & \textbf{35.9} \textcolor{gray}{$\pm$ 0.1} & \textbf{45.0} \textcolor{gray}{$\pm$ 0.2} & \textbf{77.0} & \textbf{83.4} & \textbf{69.9} & \textbf{61.1} & \textbf{77.5} & \textbf{66.3} \\
{\smaller $-$color} & \underline{16.5} \textcolor{gray}{$\pm$ 0.5} & \underline{31.4} \textcolor{gray}{$\pm$ 0.8} & \underline{38.1} \textcolor{gray}{$\pm$ 0.8} & \underline{62.0} & \underline{76.5} & \underline{69.7} & \underline{60.6} & \underline{77.4} & \underline{65.7} \\
{\smaller $-$pyramid} & \phantom{0}3.3 \textcolor{gray}{$\pm$ 0.3} & 11.0 \textcolor{gray}{$\pm$ 1.0} & 24.7 \textcolor{gray}{$\pm$ 1.0} & 41.9 & 46.4 & 69.2 & 59.7 & 77.2 & 62.1 \\
\bottomrule
\end{tabular}

\vspace{0.5em}
\caption{
\looseness=-1\textbf{Parameterization.} Our full parameterization, without color decorrelation ($-$color), and with raw pixels in place of the Laplacian pyramid ($-$pyramid). The full parameterization performs best.}
\label{tab:abl_parameterization}
\end{table}

\setlength{\intextsep}{12pt plus 2pt minus 2pt}

\clearpage
\section{More Datasets and Backbones}\label{sec:more_results}
Table~\ref{tab:from_scratch_all} extends Table~\ref{tab:from_scratch} to additional distillation backbones. Table~\ref{tab:diffusion_compare} compares against diffusion-based methods, which we exclude from the main text because they rely on a generative model pre-trained on the target data; Table~\ref{tab:featurespace_diffusion} makes the same comparison in feature space. Tables~\ref{tab:fromscratch_imagenette} and~\ref{tab:fromscratch_imagewoof} report from-scratch results on ImageNette and ImageWoof using their official splits. Finally, Table~\ref{tab:coverage_matrix} scores each of our distilled sets in every other backbone's latent space.

\begin{table}[t]
\centering
\small
\begin{tabular}{l cc cc cc}
\toprule
 & \multicolumn{2}{c}{Hard-Label} & \multicolumn{2}{c}{Soft-Label} & \multicolumn{2}{c}{Knowledge Distillation} \\
\cmidrule(lr){2-3}\cmidrule(lr){4-5}\cmidrule(lr){6-7}
Method & {\smaller IPC 10} & {\smaller IPC 50} & {\smaller IPC 10} & {\smaller IPC 50} & {\smaller IPC 10} & {\smaller IPC 50} \\
\midrule
\multicolumn{7}{l}{\emph{Noise init}} \\
ResNet-18 & 17.6 \textcolor{gray}{$\pm$ 0.2} & 18.4 \textcolor{gray}{$\pm$ 0.5} & 30.5 \textcolor{gray}{$\pm$ 0.8} & 43.1 \textcolor{gray}{$\pm$ 0.3} & \underline{44.8} \textcolor{gray}{$\pm$ 0.7} & 55.0 \textcolor{gray}{$\pm$ 0.3} \\
ViT & \underline{20.7} \textcolor{gray}{$\pm$ 0.4} & 28.4 \textcolor{gray}{$\pm$ 0.6} & 34.0 \textcolor{gray}{$\pm$ 0.6} & 48.7 \textcolor{gray}{$\pm$ 0.1} & 44.2 \textcolor{gray}{$\pm$ 0.2} & 54.9 \textcolor{gray}{$\pm$ 0.2} \\
CLIP & \phantom{0}6.1 \textcolor{gray}{$\pm$ 0.2} & \phantom{0}8.4 \textcolor{gray}{$\pm$ 0.3} & 18.5 \textcolor{gray}{$\pm$ 0.2} & 29.2 \textcolor{gray}{$\pm$ 0.5} & 32.9 \textcolor{gray}{$\pm$ 0.3} & 44.3 \textcolor{gray}{$\pm$ 0.4} \\
SigLIP2 & 14.9 \textcolor{gray}{$\pm$ 0.2} & 21.0 \textcolor{gray}{$\pm$ 0.0} & 30.7 \textcolor{gray}{$\pm$ 0.5} & 44.5 \textcolor{gray}{$\pm$ 0.7} & 40.8 \textcolor{gray}{$\pm$ 0.3} & 52.2 \textcolor{gray}{$\pm$ 0.6} \\
EVA-02 & 16.8 \textcolor{gray}{$\pm$ 0.3} & 24.7 \textcolor{gray}{$\pm$ 0.2} & 31.5 \textcolor{gray}{$\pm$ 0.3} & 45.9 \textcolor{gray}{$\pm$ 0.5} & 42.8 \textcolor{gray}{$\pm$ 0.9} & 53.8 \textcolor{gray}{$\pm$ 0.1} \\
MoCo-v3 & \textbf{27.0} \textcolor{gray}{$\pm$ 0.2} & \textbf{33.5} \textcolor{gray}{$\pm$ 1.1} & \textbf{40.3} \textcolor{gray}{$\pm$ 0.1} & \textbf{52.8} \textcolor{gray}{$\pm$ 0.0} & \textbf{48.5} \textcolor{gray}{$\pm$ 0.1} & \textbf{57.2} \textcolor{gray}{$\pm$ 0.3} \\
DINOv3 & \underline{21.3} \textcolor{gray}{$\pm$ 0.5} & \underline{30.2} \textcolor{gray}{$\pm$ 0.2} & \underline{35.9} \textcolor{gray}{$\pm$ 0.1} & \underline{50.7} \textcolor{gray}{$\pm$ 0.1} & \underline{45.0} \textcolor{gray}{$\pm$ 0.2} & \underline{56.0} \textcolor{gray}{$\pm$ 0.2} \\
\midrule
\multicolumn{7}{l}{\emph{Real init}} \\
ResNet-18 & 20.4 \textcolor{gray}{$\pm$ 0.4} & 22.0 \textcolor{gray}{$\pm$ 0.4} & 33.9 \textcolor{gray}{$\pm$ 0.2} & 45.4 \textcolor{gray}{$\pm$ 0.5} & 47.8 \textcolor{gray}{$\pm$ 0.2} & 56.4 \textcolor{gray}{$\pm$ 0.3} \\
ViT & \underline{25.6} \textcolor{gray}{$\pm$ 0.2} & \underline{37.6} \textcolor{gray}{$\pm$ 0.4} & 39.8 \textcolor{gray}{$\pm$ 0.5} & 52.9 \textcolor{gray}{$\pm$ 0.2} & 48.8 \textcolor{gray}{$\pm$ 0.3} & \underline{57.6} \textcolor{gray}{$\pm$ 0.1} \\
CLIP & 20.9 \textcolor{gray}{$\pm$ 0.2} & 36.5 \textcolor{gray}{$\pm$ 0.5} & 35.5 \textcolor{gray}{$\pm$ 0.2} & 49.1 \textcolor{gray}{$\pm$ 0.2} & 47.3 \textcolor{gray}{$\pm$ 0.2} & 55.9 \textcolor{gray}{$\pm$ 0.2} \\
SigLIP2 & 21.9 \textcolor{gray}{$\pm$ 0.2} & 32.3 \textcolor{gray}{$\pm$ 0.3} & 37.1 \textcolor{gray}{$\pm$ 0.3} & 50.0 \textcolor{gray}{$\pm$ 0.2} & 47.6 \textcolor{gray}{$\pm$ 0.1} & 56.4 \textcolor{gray}{$\pm$ 0.2} \\
EVA-02 & 24.1 \textcolor{gray}{$\pm$ 0.3} & \underline{37.7} \textcolor{gray}{$\pm$ 0.2} & 38.1 \textcolor{gray}{$\pm$ 0.1} & 51.8 \textcolor{gray}{$\pm$ 0.4} & 48.4 \textcolor{gray}{$\pm$ 0.1} & 57.1 \textcolor{gray}{$\pm$ 0.3} \\
MoCo-v3 & \textbf{28.9} \textcolor{gray}{$\pm$ 0.2} & 36.0 \textcolor{gray}{$\pm$ 0.7} & \textbf{41.7} \textcolor{gray}{$\pm$ 0.0} & \underline{53.4} \textcolor{gray}{$\pm$ 0.1} & \textbf{49.6} \textcolor{gray}{$\pm$ 0.2} & \underline{57.7} \textcolor{gray}{$\pm$ 0.2} \\
DINOv3 & \underline{26.0} \textcolor{gray}{$\pm$ 0.2} & \textbf{41.1} \textcolor{gray}{$\pm$ 0.4} & \underline{40.5} \textcolor{gray}{$\pm$ 0.0} & \textbf{54.3} \textcolor{gray}{$\pm$ 0.4} & \textbf{49.5} \textcolor{gray}{$\pm$ 0.3} & \textbf{58.1} \textcolor{gray}{$\pm$ 0.1} \\
\bottomrule
\end{tabular}

\vspace{0.5em}
\caption{
\looseness=-1
\textbf{From-scratch training across distillation backbones.} Most backbones yield distilled data that can effectively train a model from scratch, with MoCo-v3~\cite{mocov3} and DINOv3~\cite{dinov3} eking out a slight edge over the others.}
\label{tab:from_scratch_all}
\end{table}

\begin{table}[t]
\centering
\small
\begin{tabular}{l cc cc cc}
\toprule
 & \multicolumn{2}{c}{Hard-Label} & \multicolumn{2}{c}{Soft-Label} & \multicolumn{2}{c}{Knowledge Distillation} \\
\cmidrule(lr){2-3}\cmidrule(lr){4-5}\cmidrule(lr){6-7}
Method & IPC 10 & IPC 50 & IPC 10 & IPC 50 & IPC 10 & IPC 50 \\
\midrule
\multicolumn{7}{l}{\emph{Diffusion}} \\
Minimax~\cite{minimax} & 22.2 \textcolor{gray}{$\pm$ 0.2} & 36.1 \textcolor{gray}{$\pm$ 0.3} & 35.9 \textcolor{gray}{$\pm$ 0.2} & 48.8 \textcolor{gray}{$\pm$ 0.2} & 49.7 \textcolor{gray}{$\pm$ 0.1} & 58.1 \textcolor{gray}{$\pm$ 0.1} \\
D4M~\cite{d4m} & 17.0 \textcolor{gray}{$\pm$ 0.1} & 30.5 \textcolor{gray}{$\pm$ 0.2} & 35.8 \textcolor{gray}{$\pm$ 0.2} & 50.3 \textcolor{gray}{$\pm$ 0.1} & 49.6 \textcolor{gray}{$\pm$ 0.3} & 58.5 \textcolor{gray}{$\pm$ 0.1} \\
D$^3$HR~\cite{d3hr} & 23.7 \textcolor{gray}{$\pm$ 0.2} & 38.0 \textcolor{gray}{$\pm$ 0.4} & 35.7 \textcolor{gray}{$\pm$ 0.1} & 50.2 \textcolor{gray}{$\pm$ 0.2} & 48.1 \textcolor{gray}{$\pm$ 0.3} & 57.7 \textcolor{gray}{$\pm$ 0.1} \\
CaO$_2$~\cite{cao2} & \textbf{28.8} \textcolor{gray}{$\pm$ 0.2} & \textbf{44.3} \textcolor{gray}{$\pm$ 0.1} & 40.7 \textcolor{gray}{$\pm$ 0.2} & 54.1 \textcolor{gray}{$\pm$ 0.1} & \textbf{51.4} \textcolor{gray}{$\pm$ 0.3} & \underline{59.5} \textcolor{gray}{$\pm$ 0.1} \\
ManifoldGD~\cite{manifoldgd} & 27.6 \textcolor{gray}{$\pm$ 0.3} & \textbf{43.8} \textcolor{gray}{$\pm$ 0.4} & \underline{41.2} \textcolor{gray}{$\pm$ 0.2} & \textbf{54.6} \textcolor{gray}{$\pm$ 0.0} & \textbf{51.5} \textcolor{gray}{$\pm$ 0.1} & \textbf{59.8} \textcolor{gray}{$\pm$ 0.1} \\
\midrule
\multicolumn{7}{l}{\emph{Ours, noise init}} \\
ResNet-18 & 17.6 \textcolor{gray}{$\pm$ 0.2} & 18.4 \textcolor{gray}{$\pm$ 0.5} & 30.5 \textcolor{gray}{$\pm$ 0.8} & 43.1 \textcolor{gray}{$\pm$ 0.3} & 44.8 \textcolor{gray}{$\pm$ 0.7} & 55.0 \textcolor{gray}{$\pm$ 0.3} \\
MoCo-v3 & 27.0 \textcolor{gray}{$\pm$ 0.2} & 33.5 \textcolor{gray}{$\pm$ 1.1} & 40.3 \textcolor{gray}{$\pm$ 0.1} & 52.8 \textcolor{gray}{$\pm$ 0.0} & 48.5 \textcolor{gray}{$\pm$ 0.1} & 57.2 \textcolor{gray}{$\pm$ 0.3} \\
DINOv3 & 21.3 \textcolor{gray}{$\pm$ 0.5} & 30.2 \textcolor{gray}{$\pm$ 0.2} & 35.9 \textcolor{gray}{$\pm$ 0.1} & 50.7 \textcolor{gray}{$\pm$ 0.1} & 45.0 \textcolor{gray}{$\pm$ 0.2} & 56.0 \textcolor{gray}{$\pm$ 0.2} \\
\midrule
\multicolumn{7}{l}{\emph{Ours, real init}} \\
ResNet-18 & 20.4 \textcolor{gray}{$\pm$ 0.4} & 22.0 \textcolor{gray}{$\pm$ 0.4} & 33.9 \textcolor{gray}{$\pm$ 0.2} & 45.4 \textcolor{gray}{$\pm$ 0.5} & 47.8 \textcolor{gray}{$\pm$ 0.2} & 56.4 \textcolor{gray}{$\pm$ 0.3} \\
MoCo-v3 & \textbf{28.9} \textcolor{gray}{$\pm$ 0.2} & 36.0 \textcolor{gray}{$\pm$ 0.7} & \textbf{41.7} \textcolor{gray}{$\pm$ 0.0} & 53.4 \textcolor{gray}{$\pm$ 0.1} & 49.6 \textcolor{gray}{$\pm$ 0.2} & 57.7 \textcolor{gray}{$\pm$ 0.2} \\
DINOv3 & 26.0 \textcolor{gray}{$\pm$ 0.2} & 41.1 \textcolor{gray}{$\pm$ 0.4} & 40.5 \textcolor{gray}{$\pm$ 0.0} & \textbf{54.3} \textcolor{gray}{$\pm$ 0.4} & 49.5 \textcolor{gray}{$\pm$ 0.3} & 58.1 \textcolor{gray}{$\pm$ 0.1} \\
\bottomrule
\end{tabular}

\vspace{0.5em}
\caption{\looseness=-1 
\textbf{Comparison with Diffusion-Based Methods.} Our method matches the performance of the \textit{best} diffusion-based methods in the soft-label setting while falling slightly behind on hard-label and knowledge distillation. Despite not leveraging a pre-trained generative model, our distilled data remain competitive.}
\label{tab:diffusion_compare}
\end{table}

\begin{table}[t]
\centering
\small
\setlength{\tabcolsep}{3pt}%
\resizebox{\linewidth}{!}{%
\begin{tabular}{l *{4}{c} *{4}{c} *{4}{c}}
\toprule
 & \multicolumn{4}{c}{Coverage} & \multicolumn{4}{c}{$k$-NN} & \multicolumn{4}{c}{Linear} \\
\cmidrule(lr){2-5}\cmidrule(lr){6-9}\cmidrule(lr){10-13}
 & \multicolumn{2}{c}{SigLIP2} & \multicolumn{2}{c}{MoCo-v3} & \multicolumn{2}{c}{SigLIP2} & \multicolumn{2}{c}{MoCo-v3} & \multicolumn{2}{c}{SigLIP2} & \multicolumn{2}{c}{MoCo-v3} \\
\cmidrule(lr){2-3}\cmidrule(lr){4-5}\cmidrule(lr){6-7}\cmidrule(lr){8-9}\cmidrule(lr){10-11}\cmidrule(lr){12-13}
Method & {\smaller IPC 10} & {\smaller IPC 50} & {\smaller IPC 10} & {\smaller IPC 50} & {\smaller IPC 10} & {\smaller IPC 50} & {\smaller IPC 10} & {\smaller IPC 50} & {\smaller IPC 10} & {\smaller IPC 50} & {\smaller IPC 10} & {\smaller IPC 50} \\
\midrule
\multicolumn{13}{l}{\emph{Diffusion}} \\
Minimax~\cite{minimax} & 60.6 & 82.3 & 67.6 & 88.2 & 67.2 & 70.3 & 57.9 & 61.4 & 71.9 & 74.4 & 61.2 & 64.4 \\
D4M~\cite{d4m} & 37.1 & 60.3 & 47.1 & 76.3 & 57.8 & 62.3 & 48.1 & 53.5 & 65.6 & 67.7 & 55.2 & 62.2 \\
D$^3$HR~\cite{d3hr} & 55.0 & 78.9 & 65.0 & 88.5 & 67.9 & 71.0 & 59.4 & 62.7 & 72.7 & 75.1 & 62.4 & 66.2 \\
CaO$_2$~\cite{cao2} & 72.9 & 92.3 & 71.2 & 94.1 & 67.3 & 70.2 & 58.5 & 61.1 & 73.1 & 75.9 & 62.8 & 66.9 \\
ManifoldGD~\cite{manifoldgd} & \underline{77.4} & \underline{94.7} & 75.1 & 94.9 & 68.5 & 72.1 & 59.4 & 62.9 & 74.5 & 77.2 & 63.9 & 68.1 \\
\midrule
\multicolumn{13}{l}{\emph{Ours, noise init}} \\
ResNet-18 & 19.7 & 38.4 & 41.3 & 66.3 & 54.3 & 61.2 & 55.3 & 58.7 & 69.6 & 71.2 & 59.8 & 66.5 \\
MoCo-v3 & 61.2 & 86.9 & (91.9) & (97.9) & 64.3 & 68.4 & (60.2) & (62.5) & 74.6 & 74.6 & (66.6) & (69.9) \\
DINOv3 & 77.0 & 93.2 & \underline{83.4} & \underline{96.3} & \underline{69.9} & \underline{72.5} & \textbf{61.1} & \textbf{63.2} & \textbf{77.5} & \textbf{78.6} & \textbf{66.3} & \textbf{70.3} \\
\midrule
\multicolumn{13}{l}{\emph{Ours, real init}} \\
ResNet-18 & 30.0 & 57.5 & 54.9 & 82.0 & 62.1 & 68.3 & 58.0 & 61.3 & 72.8 & 74.5 & 63.1 & 67.8 \\
MoCo-v3 & 60.6 & 88.5 & (91.3) & (98.0) & 65.0 & 69.3 & (60.2) & (62.2) & 75.0 & 75.8 & (66.4) & (70.0) \\
DINOv3 & \textbf{79.9} & \textbf{95.5} & \textbf{85.1} & \textbf{97.5} & \textbf{69.9} & \textbf{72.8} & \underline{60.3} & \underline{63.0} & \underline{77.0} & \underline{78.5} & \underline{65.7} & \underline{70.2} \\
\bottomrule
\end{tabular}
}

\vspace{0.5em}
\caption{\looseness=-1 
\textbf{Feature-Space Comparison with Diffusion-Based Methods.} Unlike in the from-scratch regime, our method (DINOv3~\cite{dinov3} variant) consistently outperforms the diffusion-based methods across all feature-space metrics. This is quite remarkable given that our method does not use a generative model to constrain the distilled data to the real image manifold.}
\label{tab:featurespace_diffusion}
\end{table}

\begin{table}[t]
\centering
\small
\begin{tabular}{l cc cc}
\toprule
 & \multicolumn{2}{c}{Hard-Label} & \multicolumn{2}{c}{Knowledge Distillation} \\
\cmidrule(lr){2-3}\cmidrule(lr){4-5}
Method & IPC 10 & IPC 50 & IPC 10 & IPC 50 \\
\midrule
\multicolumn{5}{l}{\emph{Noise init}} \\
SRe\textsuperscript{2}L~\cite{sre2l} & 10.6 \textcolor{gray}{$\pm$ 0.3} & 15.9 \textcolor{gray}{$\pm$ 0.7} & 77.6 \textcolor{gray}{$\pm$ 1.0} & 91.8 \textcolor{gray}{$\pm$ 0.3} \\
CDA~\cite{cda} & 17.3 \textcolor{gray}{$\pm$ 0.8} & 32.7 \textcolor{gray}{$\pm$ 0.3} & 75.7 \textcolor{gray}{$\pm$ 2.2} & 91.0 \textcolor{gray}{$\pm$ 0.3} \\
G-VBSM~\cite{gvbsm} & 23.4 \textcolor{gray}{$\pm$ 1.2} & 21.5 \textcolor{gray}{$\pm$ 0.5} & \textbf{82.8} \textcolor{gray}{$\pm$ 0.4} & \textbf{93.6} \textcolor{gray}{$\pm$ 0.3} \\
WMDD~\cite{wmdd} & 42.4 \textcolor{gray}{$\pm$ 1.2} & 55.7 \textcolor{gray}{$\pm$ 0.1} & 81.2 \textcolor{gray}{$\pm$ 0.7} & \underline{92.4} \textcolor{gray}{$\pm$ 0.1} \\
Ours (ResNet-18) & \textbf{56.2} \textcolor{gray}{$\pm$ 0.4} & \underline{70.7} \textcolor{gray}{$\pm$ 0.7} & \textbf{82.7} \textcolor{gray}{$\pm$ 0.6} & \underline{92.4} \textcolor{gray}{$\pm$ 0.3} \\
Ours (DINOv3) & \textbf{56.4} \textcolor{gray}{$\pm$ 0.8} & \textbf{71.9} \textcolor{gray}{$\pm$ 0.3} & 79.9 \textcolor{gray}{$\pm$ 0.3} & \underline{92.2} \textcolor{gray}{$\pm$ 0.3} \\
\midrule
\multicolumn{5}{l}{\emph{Real init}} \\
RDED~\cite{rded} & 42.8 \textcolor{gray}{$\pm$ 0.8} & 64.6 \textcolor{gray}{$\pm$ 0.7} & 78.7 \textcolor{gray}{$\pm$ 0.2} & \textbf{92.8} \textcolor{gray}{$\pm$ 0.2} \\
FADRM~\cite{fadrm} & 50.4 \textcolor{gray}{$\pm$ 1.2} & 72.6 \textcolor{gray}{$\pm$ 0.5} & \textbf{83.9} \textcolor{gray}{$\pm$ 0.6} & \textbf{93.2} \textcolor{gray}{$\pm$ 0.3} \\
Ours (ResNet-18) & \textbf{60.9} \textcolor{gray}{$\pm$ 0.6} & \textbf{74.7} \textcolor{gray}{$\pm$ 1.1} & \textbf{84.5} \textcolor{gray}{$\pm$ 0.9} & \textbf{92.9} \textcolor{gray}{$\pm$ 0.4} \\
Ours (DINOv3) & \textbf{61.3} \textcolor{gray}{$\pm$ 0.6} & \textbf{75.9} \textcolor{gray}{$\pm$ 0.3} & \textbf{83.4} \textcolor{gray}{$\pm$ 1.0} & \textbf{93.1} \textcolor{gray}{$\pm$ 0.4} \\
\bottomrule
\end{tabular}

\vspace{0.5em}
\caption{\looseness=-1 
\textbf{ImageNette From-Scratch Training.} On the ``easy'' 10-class ImageNette~\cite{imagenette} dataset, we outperform the baselines in the hard-label setting and match or beat them in the knowledge distillation regime.}
\label{tab:fromscratch_imagenette}
\end{table}

\begin{table}[t]
\centering
\small
\begin{tabular}{l cc cc}
\toprule
 & \multicolumn{2}{c}{Hard-Label} & \multicolumn{2}{c}{Knowledge Distillation} \\
\cmidrule(lr){2-3}\cmidrule(lr){4-5}
Method & IPC 10 & IPC 50 & IPC 10 & IPC 50 \\
\midrule
\multicolumn{5}{l}{\emph{Noise init}} \\
SRe\textsuperscript{2}L~\cite{sre2l} & 15.3 \textcolor{gray}{$\pm$ 0.3} & 25.6 \textcolor{gray}{$\pm$ 0.5} & 56.7 \textcolor{gray}{$\pm$ 0.6} & 80.6 \textcolor{gray}{$\pm$ 0.3} \\
CDA~\cite{cda} & 15.8 \textcolor{gray}{$\pm$ 0.1} & 24.2 \textcolor{gray}{$\pm$ 0.6} & 57.1 \textcolor{gray}{$\pm$ 0.6} & 80.8 \textcolor{gray}{$\pm$ 0.2} \\
G-VBSM~\cite{gvbsm} & 18.7 \textcolor{gray}{$\pm$ 0.8} & 26.8 \textcolor{gray}{$\pm$ 0.5} & 62.3 \textcolor{gray}{$\pm$ 0.2} & \textbf{84.1} \textcolor{gray}{$\pm$ 0.3} \\
WMDD~\cite{wmdd} & 16.7 \textcolor{gray}{$\pm$ 0.2} & 20.7 \textcolor{gray}{$\pm$ 0.9} & 33.3 \textcolor{gray}{$\pm$ 0.9} & 65.9 \textcolor{gray}{$\pm$ 1.4} \\
Ours (ResNet-18) & \underline{34.2} \textcolor{gray}{$\pm$ 1.0} & \underline{53.2} \textcolor{gray}{$\pm$ 0.2} & \textbf{65.6} \textcolor{gray}{$\pm$ 0.7} & 81.7 \textcolor{gray}{$\pm$ 0.6} \\
Ours (DINOv3) & \textbf{37.7} \textcolor{gray}{$\pm$ 0.4} & \textbf{57.8} \textcolor{gray}{$\pm$ 0.8} & \textbf{66.5} \textcolor{gray}{$\pm$ 0.9} & \textbf{83.3} \textcolor{gray}{$\pm$ 0.6} \\
\midrule
\multicolumn{5}{l}{\emph{Real init}} \\
RDED~\cite{rded} & 14.6 \textcolor{gray}{$\pm$ 0.9} & 27.2 \textcolor{gray}{$\pm$ 1.2} & 60.4 \textcolor{gray}{$\pm$ 0.7} & \textbf{83.7} \textcolor{gray}{$\pm$ 0.5} \\
FADRM~\cite{fadrm} & 25.0 \textcolor{gray}{$\pm$ 0.8} & 41.6 \textcolor{gray}{$\pm$ 0.6} & 61.2 \textcolor{gray}{$\pm$ 2.2} & 81.9 \textcolor{gray}{$\pm$ 0.6} \\
Ours (ResNet-18) & \textbf{40.1} \textcolor{gray}{$\pm$ 0.8} & \underline{56.6} \textcolor{gray}{$\pm$ 1.5} & \textbf{66.3} \textcolor{gray}{$\pm$ 1.2} & 82.1 \textcolor{gray}{$\pm$ 0.4} \\
Ours (DINOv3) & \textbf{40.3} \textcolor{gray}{$\pm$ 0.7} & \textbf{59.5} \textcolor{gray}{$\pm$ 1.0} & \textbf{68.4} \textcolor{gray}{$\pm$ 1.4} & \textbf{84.3} \textcolor{gray}{$\pm$ 0.3} \\
\bottomrule
\end{tabular}

\vspace{0.5em}
\caption{\looseness=-1 
\textbf{ImageWoof From-Scratch Training.} On the ``hard'' 10-class ImageWoof~\cite{imagenette} dataset, we also outperform the baselines in the hard-label setting and match or beat them in the knowledge distillation regime.}
\label{tab:fromscratch_imagewoof}
\end{table}

\begin{table}[t]
\centering
\small
\setlength{\tabcolsep}{4pt}%
\begin{tabular}{l ccccccc}
\toprule
 & \multicolumn{7}{c}{Feature space} \\
\cmidrule(lr){2-8}
Distilled set & {\smaller ViT} & {\smaller CLIP} & {\smaller SigLIP2} & {\smaller EVA-02} & {\smaller MoCo-v3} & {\smaller EuPE} & {\smaller DINOv3} \\
\midrule
ViT~\cite{vit} & \cellcolor{blue!45}(94.6) & \cellcolor{blue!4}59.9 & \cellcolor{blue!5}50.6 & \cellcolor{blue!4}57.2 & \cellcolor{blue!36}69.9 & \cellcolor{blue!5}41.7 & \cellcolor{blue!16}39.6 \\
CLIP~\cite{clip} & \cellcolor{blue!4}29.0 & \cellcolor{blue!31}(77.2) & \cellcolor{blue!4}50.4 & \cellcolor{blue!7}59.1 & \cellcolor{blue!4}20.4 & \cellcolor{blue!4}41.3 & \cellcolor{blue!4}28.6 \\
SigLIP2~\cite{siglip2} & \cellcolor{blue!34}73.5 & \cellcolor{blue!43}84.7 & \cellcolor{blue!45}(78.0) & \cellcolor{blue!38}84.8 & \cellcolor{blue!28}56.9 & \cellcolor{blue!33}71.5 & \cellcolor{blue!33}56.1 \\
EVA-02~\cite{fang2024eva} & \cellcolor{blue!36}76.2 & \cellcolor{blue!45}86.2 & \cellcolor{blue!35}70.5 & \cellcolor{blue!45}(91.8) & \cellcolor{blue!29}58.7 & \cellcolor{blue!38}76.5 & \cellcolor{blue!42}64.6 \\
MoCo-v3~\cite{mocov3} & \cellcolor{blue!45}89.5 & \cellcolor{blue!23}72.2 & \cellcolor{blue!21}61.2 & \cellcolor{blue!27}75.7 & \cellcolor{blue!45}(91.9) & \cellcolor{blue!15}52.3 & \cellcolor{blue!25}48.2 \\
EuPE~\cite{eupe} & \cellcolor{blue!40}82.1 & \cellcolor{blue!44}85.5 & \cellcolor{blue!43}75.8 & \cellcolor{blue!45}90.4 & \cellcolor{blue!42}78.2 & \cellcolor{blue!43}(81.8) & \cellcolor{blue!45}67.3 \\
DINOv3~\cite{dinov3} & \cellcolor{blue!41}82.8 & \cellcolor{blue!45}86.1 & \cellcolor{blue!45}77.0 & \cellcolor{blue!44}89.6 & \cellcolor{blue!45}83.4 & \cellcolor{blue!45}84.1 & \cellcolor{blue!45}(95.1) \\
\bottomrule
\end{tabular}

\vspace{0.5em}
\caption{\looseness=-1 
\textbf{Inter-Model Coverage.} Coverage of our noise-initialized ImageNet-1k distilled sets (IPC 10), distilled with the backbone in each row and evaluated in the feature space of the backbone in each column. Some encoders yield data that generalize very well (e.g., DINOv3) while data yielded by other encoders seem very overfit to the source model (e.g., CLIP). Same-model performance is indicated by (parentheses).}
\label{tab:coverage_matrix}
\end{table}

\clearpage

\section{More Method Details}\label{sec:method_details}
\textbf{Note on from-scratch evaluation:} 
As pointed out in recent work~\cite{dey2026rethinking}, the most popular large-scale dataset distillation evaluation method effectively changes the task to \textit{knowledge distillation} rather than training from scratch. 
While knowledge distillation is an interesting task deserving of its own attention, we wish to focus on from-scratch training, the original goal of dataset distillation. 
For our soft-label evaluations, rather than generating new soft labels at every step, we use a teacher model to generate exactly one soft label per image before training by averaging its temperature-softened predictions over 10 random crops. 
We re-evaluate prior works under this training paradigm as well. 
\subsection*{Distillation}
For our real-initialized variants, we initialize the distilled images with those of RDED~\cite{rded} and distill for 5000 iterations. 
Our noise-initialized variants start from random noise and are distilled for 10000 iterations.
We use all real data as positive samples at every step (this is feasible even on small GPUs with $\sim$1.3k real embeddings per class). 

To optimize our distilled data, we use Adam~\cite{kingma2014adam} with a learning rate of 0.001.

Images are distilled at $256\times 256$ resolution. 
Image pyramids are initialized with the first 4 layers (up to $8\times 8$) as standard Gaussian noise. 
A new layer is added every 200 iterations (as standard Gaussian noise) until the full resolution is reached. 

\looseness=-1
Videos are distilled at $256\times 256$ spatial resolution and $16$ frames.
Video pyramids are initialized with the first 4 layers and 1 frame as standard Gaussian noise. 
A new layer with twice the spatial and temporal resolution is added every 200 iterations (as standard Gaussian noise) until the full resolution is reached. 

Audio pyramids are initialized as just the first layer. 
A new layer is added every 100 steps until the final resolution of 524,288 samples is reached.

\textbf{Drifting field implementation.} For clarity, Section~\ref{sec:drifting} presents the drifting field in its simplest form. In practice, we follow the full implementation of~\cite{drifting}: the kernel is additionally normalized over the synthetic-sample axis (their Algorithm~2), and we apply the feature and drift normalizations of their Appendix~A.6. Concretely, each feature $\phi_j$ is rescaled by a stop-gradient scale $S_j$ such that the average pairwise distance is approximately $\sqrt{C_j}$ (where $C_j$ is the feature dimension), the kernel temperature is correspondingly scaled to $\tau\sqrt{C_j}$, and the resulting field $\mathbf{V}_j$ is divided by its root-mean-square per-dimension magnitude. A single normalization scale is shared across all spatial locations of a given feature map, and the total loss is the sum of the per-feature losses.

\textbf{Features.} The feature extractor $\phi$ yields several features $\phi_j$ per sample. For ViT backbones, we use the CLS token together with all patch tokens. For ResNet-18, we use the final global-average-pooled vector together with the per-location feature vectors of the final two convolutional layers. For both audio and video methods, we drift the mean token at each layer.

\section{Evaluation Protocols}\label{sec:eval_protocols}
All from-scratch evaluations train a ResNet-18~\cite{resnet} student and report top-1 accuracy of the final-epoch model on the ImageNet-1k validation set (single $224\times224$ center crop after resizing to 256). All numbers are the mean and standard deviation over 3 training seeds. Every method, including ours, is re-evaluated with identical recipes.
\subsection{Hard-Label Training}
On ImageNet-1k~\cite{imagenet1k}, hard-label training uses the soft-label recipe below with $\alpha=0$ (i.e., only the label-smoothed cross-entropy on the one-hot labels), so the two settings differ only in the targets.

\looseness=-1
On ImageNette and ImageWoof~\cite{imagenette}, we use the evaluation protocol from \mbox{Minimax Diffusion~\cite{minimax}:}
\begin{itemize}
    \item \texttt{net\_type = resnet}
    \item \texttt{depth = 18}
    \item \texttt{norm\_type = instance}
    \item \texttt{optimizer = SGD}
    \item \texttt{lr = 0.01}
    \item \texttt{momentum = 0.9}
    \item \texttt{weight\_decay = 5e-4}
    \item \texttt{epochs (E)}
    \begin{itemize}
        \item \texttt{IPC = 1\phantom{0} \textrightarrow\;\;E = 3000}
        \item \texttt{IPC = 10 \textrightarrow\;\;E = 2000}
        \item \texttt{IPC = 50 \textrightarrow\;\;E = 1500}
    \end{itemize}
    \item \texttt{lr\_schedule = MultiStepLR}
    \begin{itemize}
        \item \texttt{milestones = [2·E\slash 3, 5·E\slash 6]}
        \item \texttt{gamma = 0.2}
    \end{itemize}
    \item \texttt{batch\_size = 64}
\end{itemize}

\subsection{Soft-Label Training}
We implement our own recipe for soft-label training and apply it equally to our method and all baselines. 
Before training, a pre-trained ResNet-18 teacher (torchvision weights) computes \textbf{exactly one soft label per distilled image}: the average of its temperature-scaled softmax outputs ($T=10$) over 10 random-resized crops (scale $[0.3, 1]$) with horizontal flips. The student is trained with 
$$\mathcal{L}=\alpha\,T^2\,\mathrm{CE}(\mathrm{softmax}(z/T), y_\text{soft}) + (1-\alpha)\,\mathrm{CE}_{\text{ls}=0.1}(z, y_\text{hard})$$
where $z$ are the student logits and $\alpha=0.9$. Other hyperparameters are as follows:
\begin{itemize}
    \item \texttt{student = resnet18 (batch norm)}
    \item \texttt{optimizer = AdamW}
    \item \texttt{lr = 5e-4}
    \item \texttt{lr\_schedule = 20-epoch linear warmup, then cosine to 1e-5}
    \item \texttt{weight\_decay = 0.5 \textrightarrow\;\;0.1 (cosine)}
    \item \texttt{epochs = 1000}
    \item \texttt{batch\_size = 50}
    \item \texttt{ema\_decay = 0.9995}
    \item \texttt{augmentation}
    \begin{itemize}
        \item \texttt{RandomResizedCrop(min\_scale=0.3)}
        \item \texttt{RandomHorizontalFlip()}
        \item \texttt{TrivialAugmentWide()}
        \item \texttt{MixUp(alpha=0.2, p=0.5)}
    \end{itemize}
\end{itemize}

\subsection{Knowledge Distillation}
Rather than from-scratch training, most existing works on large-scale dataset distillation (i.e., ImageNet-1k for ResNet-18) have evaluated using a \textbf{knowledge distillation} protocol. 
Rather than using a (hard or soft) fixed label for each distilled image, the knowledge distillation protocol uses a teacher model to recompute a new label at every training step using a newly augmented version of the image. 
For our knowledge distillation evaluations, we use the recipe defined in EDC~\cite{edc}.

The method uses a set of $M$ teacher models $f_m$ to compute a soft target for each augmented sample $x$ as the average of each model's raw output logits:
$$y_\text{soft}=\frac{1}{M}\sum_m f_m(x)$$
The total knowledge distillation loss is then
$$\mathcal{L}_\text{KD}=\mbox{MSE}(z, y_\text{soft}) + \lambda_\text{CE} \mbox{CE}(z, y_\text{hard})$$
where $z$ are the output logits of the student model, MSE is the mean-squared-error loss, CE is the cross-entropy loss, $y_\text{hard}$ is the class-specified one-hot label, and $\lambda_\text{CE}$ is the cross-entropy weight. 

Other hyperparameters are as follows:
\begin{itemize}
    \item \texttt{student = resnet18}
    \item \texttt{teachers}
    \begin{itemize}
        \item \texttt{ImageNet-1k \textrightarrow\;\;4-model EDC ensemble}
        \begin{itemize}
            \item \texttt{resnet18}
            \item \texttt{mobilenet\_v2}
            \item \texttt{shufflenet\_v2\_x0\_5}
            \item \texttt{alexnet}
        \end{itemize}
        \item \texttt{ImageNette\slash ImageWoof \textrightarrow\;\;resnet18}
    \end{itemize}
    \item \texttt{ce\_weight = 0.025}
    \item \texttt{optimizer = AdamW}
    \item \texttt{lr = 1e-3}
    \item \texttt{weight\_decay = 0.01}
    \item \texttt{lr\_schedule = cosine}
    \item \texttt{lr\_schedule\_deceleration = 2}
    \item \texttt{ema\_decay = 0.99}
    \item \texttt{augmentation}
    \begin{itemize}
        \item \texttt{RandomResizedCrop(min\_scale=0.5)}
        \item \texttt{RandomHorizontalFlip()}
        \item \texttt{RandAugment(n=2, m=9)\;\;\;(ImageNette and ImageWoof only)}
        \item \texttt{CutMix(alpha=1.0)}
    \end{itemize}
    \item \texttt{batch\_size}
    \begin{itemize}
        \item \texttt{ImageNet-1k \textrightarrow\;\;100}
        \item \texttt{ImageNette\slash ImageWoof \textrightarrow\;\;50}
    \end{itemize}
    \item \texttt{epochs (E)}
    \begin{itemize}
        \item \texttt{ImageNet-1k \textrightarrow\;\;E = 300}
        \item \texttt{ImageNette\slash ImageWoof \textrightarrow\;\;E = 1000}
    \end{itemize}
\end{itemize}

\subsection{Coverage}
For a set of distilled points, we compute the coverage~\cite{naeem2020reliable} by measuring the fraction of real \textbf{test set} points that have at least one distilled point within the ball defined by their $k$-th nearest real neighbor in the latent space of some encoder. 
We use cosine distance and $k=5$ by default. 
We compute the coverage per class and report the average. 
The pool of distilled points includes each distilled image's center crop together with 10 random-resized crops (scale $[0.08, 1]$) per image.

\subsection{$k$-Nearest Neighbors}
\looseness=-1
We compute the $k$-Nearest Neighbor ($k$-NN) performance of a distilled dataset in the typical fashion; for each real test point, we compute the $k$ nearest distilled points (in the space of some encoder) and predict the most common label among those $k$ neighbors. By default, we use cosine distance and $k=1$. As with coverage, the pool of distilled points includes each distilled image's center crop together with 10 random-resized crops (scale $[0.08, 1]$) per image.

\subsection{Linear Classification}
We train a linear classifier on frozen, unnormalized features of an augmented bank: each distilled image's center crop together with 10 random-resized crops (scale $[0.7, 1]$), sampled uniformly. We use the following parameters:
\begin{itemize}
    \item \texttt{optimizer = AdamW}
    \item \texttt{lr = 1e-3}
    \item \texttt{lr\_schedule = cosine annealing to 0 over all steps (stepped per step)}
    \item \texttt{weight\_decay = 0}
    \item \texttt{batch\_size = 256 (sampled uniformly with replacement)}
    \item \texttt{steps: IPC = 10 \textrightarrow\;\;4000, \;IPC = 50 \textrightarrow\;\;19600}
\end{itemize}

\section{Algorithms}
\algrenewcommand{\algorithmiccomment}[1]{\hfill$\triangleright$~#1}

\begin{algorithm}[H]
\caption{Distillation loss for one class (one optimization step)}
\label{alg:training_loss}
\begin{algorithmic}[1]
\REQUIRE Pyramids of the synthetic samples $\{\bx_s\}_{s=1}^{|\cs|}$, real images of the class $\{\by_k\}_{k=1}^{M}$, features $\{\phi_j\}$, augmentation distribution $P_A$, augmentations per step $n$
\STATE Decode each $\bx_s$ from its pyramid \COMMENT{Eq.~\ref{eq:pyramid}}
\STATE Sample $A_1,\dots,A_n \sim P_A$
\FOR{each feature $\phi_j$}
    \STATE $\tilde{\bx}_{a,s} \gets \phi_j(A_a(\bx_s))$ for all $a, s$ \COMMENT{$N = n|\cs|$ embeddings}
    \STATE $\tilde{\by}_k \gets \phi_j(\by_k)$ \COMMENT{un-augmented; precomputed}
    \STATE $\bv, S \gets \textsc{DriftField}\big(\texttt{stopgrad}(\tilde{\bx}), \tilde{\by}\big)$ \COMMENT{Algorithm~\ref{alg:2}}
    \STATE $\mathcal{L}_j \gets \operatorname{MSE}\big(\tilde{\bx}/S,\; \texttt{stopgrad}(\tilde{\bx}/S + \bv)\big)$
\ENDFOR
\STATE \textbf{return} $\mathcal{L} = \sum_j \mathcal{L}_j$ \COMMENT{back-propagated to the pyramids}
\end{algorithmic}\label{alg:1}
\end{algorithm}

\begin{algorithm}[H]
\caption{\textsc{DriftField}: the drifting field for one feature $\phi_j$, following Alg.~2 and App.~A.6 of~\cite{drifting}. All steps are applied independently at each spatial location of $\phi_j$; only $S$ and $\lambda_\tau$ are computed jointly over all locations.}
\label{alg:compute_V}
\begin{algorithmic}[1]
\REQUIRE Synthetic embeddings $\{\tilde{\bx}_i\}_{i=1}^{N}$ (detached), real embeddings $\{\tilde{\by}_k\}_{k=1}^{M}$, feature dimension $C$, temperatures $\mathcal{T}$
\ENSURE Normalized drifting field $\bv(\tilde{\bx}_i)$ for $i=1,\dots,N$, feature scale $S$
\STATE $d^+_{ik} \gets \|\tilde{\bx}_i - \tilde{\by}_k\|_2$, \;\; $d^-_{il} \gets \|\tilde{\bx}_i - \tilde{\bx}_l\|_2$
\STATE $S \gets \texttt{stopgrad}\big(\operatorname{mean}(\{d^+_{ik}\} \cup \{d^-_{il}\})\big)/\sqrt{C}$ \COMMENT{feature normalization}
\STATE Divide all $d$, $\tilde{\bx}$, and $\tilde{\by}$ by $S$
\STATE $d^-_{ii} \gets \infty$ \COMMENT{mask self-pairs}
\STATE $\bv(\tilde{\bx}_i) \gets \mathbf{0}$
\FOR{$\tau \in \mathcal{T}$}
    \STATE $\ell_{i,\cdot} \gets -\big[\,d^+_{i,1:M} \,\|\, d^-_{i,1:N}\,\big] / (\tau\sqrt{C})$ \COMMENT{logits over all targets}
    \STATE $A \gets \sqrt{\operatorname{softmax}_{\text{targets}}(\ell) \odot \operatorname{softmax}_{i}(\ell)}$ \COMMENT{normalize over both axes}
    \STATE Split $A = [A^+ \,\|\, A^-]$ into positive and negative columns
    \STATE $W^+_{ik} \gets A^+_{ik}\sum_{l} A^-_{il}$, \;\; $W^-_{il} \gets A^-_{il}\sum_{k} A^+_{ik}$
    \STATE $\bv_\tau(\tilde{\bx}_i) \gets \sum_{k} W^+_{ik}\,\tilde{\by}_k - \sum_{l} W^-_{il}\,\tilde{\bx}_l$
    \STATE $\lambda_\tau \gets \sqrt{\operatorname{mean}_i \|\bv_\tau(\tilde{\bx}_i)\|_2^2 / C}$ \COMMENT{drift normalization}
    \STATE $\bv(\tilde{\bx}_i) \gets \bv(\tilde{\bx}_i) + \bv_\tau(\tilde{\bx}_i)/\lambda_\tau$
\ENDFOR
\STATE \textbf{return} $\{\bv(\tilde{\bx}_i)\}_{i=1}^{N}$, $S$
\end{algorithmic}\label{alg:2}
\end{algorithm}

\section{Limitations, Compute, Broader Impacts, and Licenses}\label{sec:limitations}
\subsection*{Limitations}
Our largest limitation is the time needed to distill the data. 
For 10 images per class, each class takes $\sim$0.25 GPU hours (more or less depending on GPU model), corresponding to roughly 250 GPU hours for ImageNet-1k before accounting for hyperparameter sweeps, multiple backbones, and larger IPC settings. 
Although distilling each class separately makes the process easy to parallelize, the total compute cost is significant.

\subsection*{Compute Resources}
We ran our experiments on a shared cluster using whatever GPUs we could get. 
Used GPUs include RTX 3090s, RTX 4090s, L40Ss, V100s, A100s, H100s, and H200s. 
All image experiments could comfortably fit on a 24 GB GPU.
Audio and video experiments required an A100, H100, or H200.
\subsection*{Broader Impacts}
Our method does not come with any readily apparent negative effects. 
Advances in dataset distillation may, in fact, lead to more efficient training, thereby shrinking the AI industry's carbon footprint. 
\subsection*{Licenses}
\paragraph*{Datasets.}

\looseness=-1
For ImageNet~\citep{imagenet1k} we build on the Torchvision~\citep{torchvision2016} implementation as part of the PyTorch~\citep{pytorch,pytorch2} ecosystem. 
ImageNet has a unique non-commercial \href{https://image-net.org/download.php}{license}, to which we have agreed. 
The remaining datasets are used as follows:
\begin{itemize}
    \item \looseness=-1\textbf{ImageNette and ImageWoof}~\cite{imagenette}: the official \texttt{imagenette2} and \texttt{imagewoof2} releases (Apache-2.0); both are subsets of ImageNet and are also subject to the ImageNet terms above.
    \item \textbf{FMA}~\cite{fma_dataset}: the \texttt{fma\_small} subset. Its metadata is released under CC BY 4.0, and each audio track is distributed under the Creative Commons license chosen by its artist; the dataset is intended for research use.
    \item \textbf{HMDB51}~\cite{kuehne2011hmdb}: released under CC BY 4.0.
    \item \textbf{Paintings}: the Claude Monet and Qi Baishi paintings in Section~\ref{sec:art} were collected from WikiArt, the Art Institute of Chicago, and the Metropolitan Museum of Art. Images from the Art Institute of Chicago and the Met are public-domain works released under CC0; WikiArt presents public-domain works and, under a fair-use rationale, copyrighted works. Monet's paintings are in the public domain. We use all paintings solely for non-commercial research.
\end{itemize}

\paragraph*{Models.}
\looseness=-1
All backbones are used with their publicly released pre-trained weights and are kept frozen.
\begin{itemize}
    \item \textbf{ResNet-18}~\cite{resnet} and \textbf{ViT-B/16}~\cite{vit}: Torchvision~\cite{torchvision2016} \texttt{IMAGENET1K\_V1} weights (BSD-3-Clause). The ResNet-18, MobileNetV2, ShuffleNetV2, and AlexNet teachers used for soft labels and knowledge distillation also come from Torchvision (BSD-3-Clause).
    \item \textbf{CLIP}~\cite{clip}: \href{https://huggingface.co/openai/clip-vit-base-patch16}{\texttt{openai/clip-vit-base-patch16}} on Hugging Face~\cite{huggingface2016}, released by OpenAI under the MIT license.
    \item \textbf{SigLIP2}~\cite{siglip2}: \href{https://huggingface.co/google/siglip2-base-patch16-224}{\texttt{google/siglip2-base-patch16-224}} (Apache-2.0).
    \item \textbf{EVA-02}~\cite{fang2024eva}: \href{https://huggingface.co/timm/eva02_base_patch14_224.mim_in22k}{\texttt{eva02\_base\_patch14\_224.mim\_in22k}} from PyTorch Image Models~\cite{rw2019timm} (MIT).
    \item \textbf{MoCo-v3}~\cite{mocov3}: ViT-B weights loaded from \href{https://huggingface.co/nyu-visionx/moco-v3-vit-b}{\texttt{nyu-visionx/moco-v3-vit-b}}; the official MoCo-v3 release is licensed under CC BY-NC 4.0.
    \item \textbf{EuPE}~\cite{eupe}: \href{https://huggingface.co/facebook/EUPE-ViT-B}{\texttt{facebook/EUPE-ViT-B}} (FAIR Noncommercial Research License).
    \item \textbf{DINOv3}~\cite{dinov3}: \href{https://huggingface.co/facebook/dinov3-vitb16-pretrain-lvd1689m}{\texttt{facebook/dinov3-vitb16-pretrain-lvd1689m}}, along with the ViT-S/16 and ViT-L/16 variants for the model-size ablation (\href{https://ai.meta.com/resources/models-and-libraries/dinov3-license}{DINOv3 License}).
    \item \textbf{VideoMAE-v2}~\cite{wang2023videomaev2}: \href{https://huggingface.co/OpenGVLab/VideoMAEv2-Base}{\texttt{OpenGVLab/VideoMAEv2-Base}} (CC BY-NC 4.0).
    \item \textbf{MusicFM}~\cite{won2023musicfm}: the FMA checkpoint from \href{https://huggingface.co/minzwon/MusicFM}{\texttt{minzwon/MusicFM}} (MIT).
\end{itemize}

\clearpage

\end{document}